\documentclass[twoside,11pt]{article}
\pdfoutput=1

\usepackage{amsmath}
\usepackage{fancyvrb}
\usepackage[utf8]{inputenc}
\usepackage{csquotes}
\usepackage{array}
\usepackage{multirow}
\usepackage{tabularx}
\usepackage{colortbl}
\usepackage{hhline}
\usepackage{subcaption}
\usepackage{lscape}
\usepackage[figuresright]{rotating}
\usepackage{ctable} 
\usepackage{float}
\usepackage{color}
\usepackage{listings}
\usepackage{jmlr2e}
\usepackage{hyperref}
\usepackage[bottom]{footmisc}
\usepackage{lastpage}

\if@usehyper
\usepackage[colorlinks=false,allbordercolors={1 1 1}]{hyperref}
\fi

\definecolor{codegreen}{rgb}{0,0.6,0}
\definecolor{codegray}{rgb}{0.5,0.5,0.5}
\definecolor{codepurple}{rgb}{0.58,0,0.82}
\definecolor{backcolour}{rgb}{0.95,0.95,0.92}
\lstdefinestyle{mystyle}{
  basicstyle=\footnotesize,
  numberstyle=\tiny\color{codegray},
  numbers=left,
  numbersep=5pt,
  xleftmargin=.1\textwidth,
  xrightmargin=.1\textwidth,
  frame=none,
  breaklines=false
}
\long\def\acks#1{\vskip 0.3in\noindent{\large\bf Acknowledgements}\vskip 0.2in
\noindent #1}

\VerbatimFootnotes

\newcommand{\relative}{\mathbin{\|}}

\newcommand\MyBox[2]{
  \fbox{\lower0.75cm
    \vbox to 1.7cm{\vfil
      \hbox to 1.7cm{\hfil\parbox{1.4cm}{#1\\#2}\hfil}
      \vfil}%
  }%
}

\DeclareMathOperator*{\argmax}{arg\,max}
\DeclareMathOperator*{\argmin}{arg\,min}

\jmlrheading{19}{2018}{1-\pageref{LastPage}}{9/16; Revised 7/18}{9/18}{16-474}{Timothy C. Au}

\ShortHeadings{The ``Absent Levels'' Problem}{Au}

\firstpageno{1}

\begin{document}
\title{Random Forests, Decision Trees, and Categorical Predictors: The ``Absent Levels'' Problem}

\author{\name Timothy C. Au \email timau@google.com\\
       \addr Google LLC \\
       1600 Amphitheatre Parkway\\
       Mountain View, CA 94043, USA}

\editor{Sebastian Nowozin}

\maketitle

\begin{abstract}
One advantage of decision tree based methods like random forests is their
ability to natively handle categorical predictors without having to first
transform them (e.g., by using feature engineering techniques).  However, in
this paper, we show how this capability can lead to an inherent ``absent
levels'' problem for decision tree based methods that has never been thoroughly
discussed, and whose consequences have never been carefully explored.  This
problem occurs whenever there is an indeterminacy over how to handle an
observation that has reached a categorical split which was determined when the
observation in question's level was absent during training.  Although these
incidents may appear to be innocuous, by using Leo Breiman and Adele Cutler's
random forests \texttt{FORTRAN} code and the \texttt{randomForest} \texttt{R}
package \citep{liaw:wiener:2002} as motivating case studies, we examine how
overlooking the absent levels problem can systematically bias a model.
Furthermore, by using three real data examples, we illustrate how absent levels
can dramatically alter a model's performance in practice, and we empirically
demonstrate how some simple heuristics can be used to help mitigate the effects
of the absent levels problem until a more robust theoretical solution is found.
\end{abstract}


\section{Introduction}
\label{sec:intro}

Since its introduction in \citet{breiman:2001}, random forests have enjoyed much
success as one of the most widely used decision tree based methods in machine
learning.  But despite their popularity and apparent simplicity, random forests
have proven to be very difficult to analyze.  Indeed, many of the basic
mathematical properties of the algorithm are still not completely well
understood, and theoretical investigations have often had to rely on either
making simplifying assumptions or considering variations of the standard
framework in order to make the analysis more tractable---see, for example,
\citet{biau:etal:2008}, \citet{biau:2012}, and \citet{denil:etal:2014}.

One advantage of decision tree based methods like random forests is their
ability to natively handle categorical predictors without having to first
transform them (e.g., by using feature engineering techniques).  However, in
this paper, we show how this capability can lead to an inherent ``absent
levels'' problem for decision tree based methods that has, to the best of our
knowledge, never been thoroughly discussed, and whose consequences have never
been carefully explored.  This problem occurs whenever there is an indeterminacy
over how to handle an observation that has reached a categorical split which was
determined when the observation in question's level was absent during
training---an issue that can arise in three different ways:
\begin{enumerate}
\item The levels are present in the population but, due to sampling variability,
  are absent in the training set.
\item The levels are present in the training set but, due to bagging, are absent
  in an individual tree's bootstrapped sample of the training set.
\item The levels are present in an individual tree's training set but, due to a
  series of earlier node splits, are absent in certain branches of the tree.
\end{enumerate}
These occurrences subsequently result in situations where observations with
absent levels are unsure of how to proceed further down the tree---an intrinsic
problem for decision tree based methods that has seemingly been overlooked in
both the theoretical literature and in much of the software that implements
these methods.

Although these incidents may appear to be innocuous, by using Leo Breiman and
Adele Cutler's random forests \texttt{FORTRAN} code and the
\texttt{randomForest} \texttt{R} package \citep{liaw:wiener:2002} as motivating
case studies,\footnote{Breiman and Cutler's random forests FORTRAN code is
  available online at: \\ \indent\indent\indent\indent
  \url{https://www.stat.berkeley.edu/~breiman/RandomForests/}} we examine how
overlooking the absent levels problem can systematically bias a model.  In
addition, by using three real data examples, we illustrate how absent levels can
dramatically alter a model's performance in practice, and we empirically
demonstrate how some simple heuristics can be used to help mitigate their
effects.

The rest of this paper is organized as follows.  In
Section~\ref{sec:background}, we introduce some notation and provide an overview
of the random forests algorithm.  Then, in Section~\ref{sec:absent}, we use
Breiman and Cutler's random forests \texttt{FORTRAN} code and the
\texttt{randomForest} \texttt{R} package to motivate our investigations into the
potential issues that can emerge when the absent levels problem is overlooked.
And although a comprehensive theoretical analysis of the absent levels problem
is beyond the scope of this paper, in Section~\ref{sec:heuristics}, we consider
some simple heuristics which may be able to help mitigate its effects.
Afterwards, in Section~\ref{sec:examples}, we present three real data examples
that demonstrate how the treatment of absent levels can significantly influence
a model's performance in practice.  Finally, we offer some concluding remarks in
Section~\ref{sec:conclusion}.

\section{Background}
\label{sec:background}

In this section, we introduce some notation and provide an overview of the
random forests algorithm.  Consequently, the more knowledgeable reader may only
need to review Sections~\ref{sec:ordered_split} and \ref{sec:unorderd_split}
which cover how the algorithm's node splits are determined.

\subsection{Classification and Regression Trees (CART)}
\label{sec:cart}

We begin by discussing the Classification and Regression Trees (CART)
methodology since the random forests algorithm uses a slightly modified version
of CART to construct the individual decision trees that are used in its
ensemble.  For a more complete overview of CART, we refer the reader to
\citet{breiman:etal:1984} or \citet{hastie:etal:2009}.

Suppose that we have a training set with $N$ independent observations
\[
\left(x_{n}, y_{n}\right), \quad n = 1, 2, \ldots, N,
\]
where $x_{n} = \left(x_{n1}, x_{n2}, \ldots, x_{nP}\right)$ and $y_{n}$ denote,
respectively, the $P$-dimensional feature vector and response for observation
$n$.  Given this initial training set, CART is a greedy recursive binary
partitioning algorithm that repeatedly partitions a larger subset of the
training set $\mathcal{N}_{\mathcal{M}} \subseteq \left\{1, 2, \ldots,
N\right\}$ (the ``mother node'') into two smaller subsets
$\mathcal{N}_{\mathcal{L}}$ and $\mathcal{N}_{\mathcal{R}}$ (the ``left'' and
``right'' daughter nodes, respectively).  Each iteration of this splitting
process, which can be referred to as ``growing the tree,'' is accomplished by
determining a decision rule that is characterized by a ``splitting variable'' $p
\in \left\{1, 2, \ldots, P\right\}$ and an accompanying ``splitting criterion''
set $\mathcal{S}_{p}$ which defines the subset of predictor $p$'s domain that
gets sent to the left daughter node $\mathcal{N}_{\mathcal{L}}$.  In particular,
any splitting variable and splitting criterion pair $\left(p,
\mathcal{S}_{p}\right)$ will partition the mother node
$\mathcal{N}_{\mathcal{M}}$ into the left and right daughter nodes which are
defined, respectively, as
\begin{equation}
\mathcal{N}_{\mathcal{L}}\!\left(p, \mathcal{S}_{p}\right) = \left\{n \in \mathcal{N}_{\mathcal{M}} : x_{np} \in \mathcal{S}_{p}\right\} \quad\mbox{and}\quad
\mathcal{N}_{\mathcal{R}}\!\left(p, \mathcal{S}_{p}\right) = \left\{n \in \mathcal{N}_{\mathcal{M}} : x_{np} \in \mathcal{S}_{p}'\right\},
\label{eq:daughter_regions}
\end{equation}
where $\mathcal{S}_{p}'$ denotes the complement of the splitting criterion set
$\mathcal{S}_{p}$ with respect to predictor $p$'s domain.  A simple model useful
for making predictions and inferences is then subsequently fit to the subset of
the training data that is in each node.

This recursive binary partitioning procedure is continued until some stopping
rule is reached---a tuning parameter that can be controlled, for example, by
placing a constraint on the minimum number of training observations that are
required in each node.  Afterwards, to help guard against overfitting, the tree
can then be ``pruned''---although we will not discuss this further as pruning
has not traditionally been done in the trees that are grown in random forests
\citep{breiman:2001}.  Predictions and inferences can then be made on an
observation by first sending it down the tree according to the tree's set of
decision rules, and then by considering the model that was fit in the furthest
node of the tree that the observation is able to reach.

The CART algorithm will grow a tree by selecting, from amongst all possible
splitting variable and splitting criterion pairs $\left(p,
\mathcal{S}_{p}\right)$, the optimal pair $\left(p^{*},
\mathcal{S}_{p^{*}}^{*}\!\right)$ which minimizes some measure of ``node
impurity'' in the resulting left and right daughter nodes as defined in
\eqref{eq:daughter_regions}.  However, the specific node impurity measure that
is being minimized will depend on whether the tree is being used for regression
or classification.

In a regression tree, the responses in a node $\mathcal{N}$ are modeled using a
constant which, under a squared error loss, is estimated by the mean of the
training responses that are in the node---a quantity which we denote as:
\begin{equation}
\hat{c}\!\left(\mathcal{N}\right) = \text{ave}\!\left(y_{n} \mid n \in \mathcal{N}\right).
\label{eq:reg_pred}
\end{equation}
Therefore, the CART algorithm will grow a regression tree by partitioning a
mother node $\mathcal{N}_{\mathcal{M}}$ on the splitting variable and splitting
criterion pair $\left(p^{*}, \mathcal{S}_{p^{*}}^{*}\!\right)$ which minimizes
the squared error resulting from the two daughter nodes that are created with
respect to a $\left(p, \mathcal{S}_{p}\right)$ pair:
\begin{equation}
  \!\left(p^{*}, \mathcal{S}_{p^{*}}^{*}\!\right) = \argmin_{\!\left(p, \mathcal{S}_{p}\right)}\!\left({\sum_{n \in \mathcal{N}_{\mathcal{L}}\!\left(p, \mathcal{S}_{p}\right)}{\!\left[y_{n} - \hat{c}\!\left(\mathcal{N}_{\mathcal{L}}\!\left(p,\mathcal{S}_{p}\right)\right)\right]^{2}} + \sum_{n \in \mathcal{N}_{\mathcal{R}}\!\left(p, \mathcal{S}_{p}\right)}{\!\left[y_{n} - \hat{c}\!\left(\mathcal{N}_{\mathcal{R}}\!\left(p,\mathcal{S}_{p}\right)\right)\right]^{2}}}\right),
\label{eq:daughter_sse}
\end{equation}
where the nodes $\mathcal{N}_{\mathcal{L}}\!\left(p, \mathcal{S}_{p}\right)$ and
$\mathcal{N}_{\mathcal{R}}\!\left(p, \mathcal{S}_{p}\right)$ are as defined in
\eqref{eq:daughter_regions}.

Meanwhile, in a classification tree where the response is categorical with $K$
possible response classes which are indexed by the set $\mathcal{K} = \left\{1,
2, \ldots, K\right\}$, we denote the proportion of training observations that
are in a node $\mathcal{N}$ belonging to each response class $k$ as:
\[
\hat{\pi}_{k}\!\left(\mathcal{N}\right) = \frac{1}{\left|\mathcal{N}\right|}\sum_{n \in \mathcal{N}}{I\!\left(y_{n} = k\right)}, \quad k \in \mathcal{K},
\]
where $\left|\cdot\right|$ is the set cardinality function and
$I\!\left(\cdot\right)$ is the indicator function.  Node $\mathcal{N}$ will then
classify its observations to the majority response class
\begin{equation}
\hat{k}\!\left(\mathcal{N}\right) = \argmax_{k \in \mathcal{K}}\,\hat{\pi}_{k}\!\left(\mathcal{N}\right),
\label{eq:class_pred}
\end{equation}
with the Gini index
\[
G\!\left(\mathcal{N}\right) = \sum_{k = 1}^{K}{\left[\hat{\pi}_{k}\!\left(\mathcal{N}\right) \cdot \left(1 - \hat{\pi}_{k}\!\left(\mathcal{N}\right)\right)\right]}
\]
providing one popular way of quantifying the node impurity in $\mathcal{N}$.
Consequently, the CART algorithm will grow a classification tree by partitioning
a mother node $\mathcal{N}_{\mathcal{M}}$ on the splitting variable and
splitting criterion pair $\left(p^{*}, \mathcal{S}_{p^{*}}^{*}\!\right)$ which
minimizes the weighted Gini index resulting from the two daughter nodes that are
created with respect to a $\left(p, \mathcal{S}_{p}\right)$ pair:
\begin{equation}
\left(p^{*}, \mathcal{S}_{p^{*}}^{*}\!\right) = \argmin_{\left(p, \mathcal{S}_{p}\right)}\!\left(\frac{{\left|\mathcal{N}_{\mathcal{L}}\!\left(p, \mathcal{S}_{p}\right)\right| \cdot G\!\left(\mathcal{N}_{\mathcal{L}}\!\left(p, \mathcal{S}_{p}\right)\right) + \left|\mathcal{N}_{\mathcal{R}}\!\left(p, \mathcal{S}_{p}\right)\right| \cdot G\!\left(\mathcal{N}_{\mathcal{R}}\!\left(p, \mathcal{S}_{p}\right)\right)}}{\left|\mathcal{N}_{\mathcal{L}}\!\left(p, \mathcal{S}_{p}\right)\right| + \left|\mathcal{N}_{\mathcal{R}}\!\left(p, \mathcal{S}_{p}\right)\right|}\right),
\label{eq:daughter_gini}
\end{equation}
where the nodes $\mathcal{N}_{\mathcal{L}}\!\left(p, \mathcal{S}_{p}\right)$ and
$\mathcal{N}_{\mathcal{R}}\!\left(p, \mathcal{S}_{p}\right)$ are as defined in
\eqref{eq:daughter_regions}.

Therefore, the CART algorithm will grow both regression and classification trees
by partitioning a mother node $\mathcal{N}_{\mathcal{M}}$ on the splitting
variable and splitting criterion pair $(p^{*}, \mathcal{S}_{p^*}^{*})$ which
minimizes the requisite node impurity measure across all possible $\left(p,
\mathcal{S}_{p}\right)$ pairs---a task which can be accomplished by first
determining the optimal splitting criterion $\mathcal{S}_{p}^{*}$ for every
predictor $p \in \left\{1, 2, \ldots, P\right\}$.  However, the specific manner
in which any particular predictor $p$'s optimal splitting criterion
$\mathcal{S}_{p}^{*}$ is determined will depend on whether $p$ is an ordered or
categorical predictor.

\subsubsection{Splitting on an Ordered Predictor}
\label{sec:ordered_split}

The splitting criterion $\mathcal{S}_{p}$ for an ordered predictor $p$ is
characterized by a numeric ``split point'' $s_{p} \in \mathbb{R}$ that defines
the half-line $\mathcal{S}_{p} = \left\{x \in \mathbb{R} : x \leq
s_{p}\right\}$.  Thus, as can be observed from \eqref{eq:daughter_regions}, a
$\left(p, \mathcal{S}_{p}\right)$ pair will partition a mother node
$\mathcal{N}_{\mathcal{M}}$ into the left and right daughter nodes that are
defined, respectively, by
\[
\mathcal{N}_{\mathcal{L}}\!\left(p, \mathcal{S}_{p}\right) = \left\{n \in \mathcal{N}_{\mathcal{M}} : x_{np} \leq s_{p}\right\} \quad\text{and}\quad
\mathcal{N}_{\mathcal{R}}\!\left(p, \mathcal{S}_{p}\right) = \left\{n \in \mathcal{N}_{\mathcal{M}} : x_{np} > s_{p}\right\}.
\]
Therefore, determining the optimal splitting criterion $\mathcal{S}_{p}^{*} =
\left\{x \in \mathbb{R} : x \leq s_{p}^{*}\right\}$ for an ordered predictor
$p$ is straightforward---it can be greedily found by searching through all of
the observed training values in the mother node in order to find the optimal
numeric split point $s_{p}^{*} \in \left\{x_{np} \in \mathbb{R} : n \in
\mathcal{N}_{\mathcal{M}}\right\}$ that minimizes the requisite node impurity
measure which is given by either \eqref{eq:daughter_sse} or
\eqref{eq:daughter_gini}.

\subsubsection{Splitting on a Categorical Predictor}
\label{sec:unorderd_split}

For a categorical predictor $p$ with $Q$ possible unordered levels which are
indexed by the set $\mathcal{Q} = \left\{1, 2, \ldots, Q\right\}$, the splitting
criterion $\mathcal{S}_{p} \subset \mathcal{Q}$ is defined by the subset of
levels that gets sent to the left daughter node $\mathcal{N}_{\mathcal{L}}$,
while the complement set $\mathcal{S}_{p}' = \mathcal{Q} \setminus
\mathcal{S}_{p}$ defines the subset of levels that gets sent to the right
daughter node $\mathcal{N}_{\mathcal{R}}$.  For notational simplicity and ease
of exposition, in the remainder of this section we assume that all $Q$ unordered
levels of $p$ are present in the mother node $\mathcal{N}_{\mathcal{M}}$ during
training since it is only these present levels which contribute to the measure
of node impurity when determining $p$'s optimal splitting criterion
$\mathcal{S}_{p}^{*}$.  Later, in Section~\ref{sec:absent}, we extend our
notation to also account for any unordered levels of a categorical predictor $p$
which are absent from the mother node $\mathcal{N}_{\mathcal{M}}$ during
training.

Consequently, there are are $2^{Q - 1} - 1$ non-redundant ways of partitioning
the $Q$ unordered levels of $p$ into the two daughter nodes, making it
computationally expensive to evaluate the resulting measure of node impurity for
every possible split when $Q$ is large.  However, this computation simplifies in
certain situations.

In the case of a regression tree with a squared error node impurity measure, a
categorical predictor $p$'s optimal splitting criterion $\mathcal{S}_{p}^{*}$
can be determined by using a procedure described in \citet{fisher:1958}.
Specifically, the training observations in the mother node are first used to
calculate the mean response within each of $p$'s unordered levels:
\begin{equation}
  \gamma_{p}(q) = \text{ave}\!\left(y_{n} \mid n \in \mathcal{N}_{\mathcal{M}} \text{ and } x_{np} = q\right), \quad q \in \mathcal{Q}.
  \label{eq:level_means}
\end{equation}
These means are then used to assign numeric ``pseudo values'' $\tilde{x}_{np}
\in \mathbb{R}$ to every training observation that is in the mother node
according to its observed level for predictor $p$:
\begin{equation}
\tilde{x}_{np} = \gamma_{p}(x_{np}), \quad n \in \mathcal{N}_{\mathcal{M}}.
\label{eq:pseudo_x}
\end{equation}
Finally, the optimal splitting criterion $\mathcal{S}_{p}^{*}$ for the
categorical predictor $p$ is determined by doing an ordered split on these
numeric pseudo values $\tilde{x}_{np}$---that is, a corresponding optimal
``pseudo splitting criterion'' $\tilde{\mathcal{S}}_{p}^{*} = \left\{\tilde{x}
\in \mathbb{R} : \tilde{x} \leq \tilde{s}_{p}^{*}\right\}$ is greedily chosen by
scanning through all of the assigned numeric pseudo values in the mother node in
order to find the optimal numeric ``pseudo split point'' $\tilde{s}_{p}^{*} \in
\left\{\tilde{x}_{np} \in \mathbb{R} : n \in \mathcal{N}_{\mathcal{M}}\right\}$
which minimizes the resulting squared error node impurity measure given in
\eqref{eq:daughter_sse} with respect to the left and right daughter nodes that
are defined, respectively, by
\begin{equation}
\mathcal{N}_{\mathcal{L}}(p, \tilde{\mathcal{S}}_{p}^{*}) = \left\{n \in \mathcal{N}_{\mathcal{M}} : \tilde{x}_{np} \leq \tilde{s}_{p}^{*}\right\} \quad\text{and}\quad
\mathcal{N}_{\mathcal{R}}(p, \tilde{\mathcal{S}}_{p}^{*}) = \left\{n \in \mathcal{N}_{\mathcal{M}} : \tilde{x}_{np} > \tilde{s}_{p}^{*}\right\}.
\label{eq:unordered_daughters_pseudo}
\end{equation}

Meanwhile, in the case of a classification tree with a weighted Gini index node
impurity measure, whether the computation simplifies or not is dependent on the
number of response classes.  For the $K > 2$ multiclass classification context,
no such simplification is possible, although several approximations have been
proposed \citep{loh:vanichsetaku:1988}.  However, for the $K = 2$ binary
classification situation, a similar procedure to the one that was just described
for regression trees can be used.  Specifically, the proportion of the training
observations in the mother node that belong to the $k = 1$ response class is
first calculated within each of categorical predictor $p$'s unordered levels:
\begin{equation}
  \gamma_{p}(q) = \frac{\left|\left\{n \in \mathcal{N}_{\mathcal{M}} : x_{np} = q \text{ and } y_{n} = 1\right\}\right|}{\left|\left\{n \in \mathcal{N}_{\mathcal{M}} : x_{np} = q\right\}\right|}, \quad q \in \mathcal{Q},
  \label{eq:level_props}
\end{equation}
and where we note here that $\gamma_{p}(q) \geq 0$ for all $q$ since these
proportions are, by definition, nonnegative.  Afterwards, and just as in
equation~\eqref{eq:pseudo_x}, these $k = 1$ response class proportions are used
to assign numeric pseudo values $\tilde{x}_{np} \in \mathbb{R}$ to every
training observation that is in the mother node $\mathcal{N}_{\mathcal{M}}$
according to its observed level for predictor $p$.  And once again, the optimal
splitting criterion $\mathcal{S}_{p}^{*}$ for the categorical predictor $p$ is
then determined by performing an ordered split on these numeric pseudo values
$\tilde{x}_{np}$---that is, a corresponding optimal pseudo splitting criterion
$\tilde{\mathcal{S}}_{p}^{*} = \left\{x \in \mathbb{R} : x \leq
\tilde{s}_{p}^{*}\right\}$ is greedily found by searching through all of the
assigned numeric pseudo values in the mother node in order to find the optimal
numeric pseudo split point $\tilde{s}_{p}^{*} \in \left\{\tilde{x}_{np} \in
\mathbb{R} : n \in \mathcal{N}_{\mathcal{M}}\right\}$ which minimizes the
weighted Gini index node impurity measure given by \eqref{eq:daughter_gini} with
respect to the resulting two daughter nodes as defined in
\eqref{eq:unordered_daughters_pseudo}.  The proof that this procedure gives the
optimal split in a binary classification tree in terms of the weighted Gini
index amongst all possible splits can be found in \citet{breiman:etal:1984} and
\citet{ripley:1996}.

Therefore, in both regression and binary classification trees, we note that the
optimal splitting criterion $\mathcal{S}_{p}^{*}$ for a categorical predictor
$p$ can be expressed in terms of the criterion's associated optimal numeric
pseudo split point $\tilde{s}_{p}^{*}$ and the requisite means or $k = 1$
response class proportions $\gamma_{p}(q)$ of the unordered levels $q \in
\mathcal{Q}$ of $p$ as follows:
\begin{itemize}
\item The unordered levels of $p$ that are being sent
  \textit{left} have means or $k = 1$ response class proportions $\gamma_{p}(q)$
  that are \textit{less than or equal to} $\tilde{s}_{p}^{*}$:
\begin{equation}
\mathcal{S}_{p}^{*} = \left\{q \in \mathcal{Q} : \gamma_{p}(q) \leq \tilde{s}_{p}^{*}\right\}.
\label{eq:levels_left}
\end{equation}
\item The unordered levels of $p$ that are being sent \textit{right} have means
  or $k = 1$ response class proportions $\gamma_{p}(q)$ that are \textit{greater
    than} $\tilde{s}_{p}^{*}$:
\begin{equation}
{\mathcal{S}_{p}^{*}}' = \left\{q \in \mathcal{Q} : \gamma_{p}(q) > \tilde{s}_{p}^{*}\right\}.
\label{eq:levels_right}
\end{equation}
\end{itemize}
As we later discuss in Section~\ref{sec:absent},
equations~\eqref{eq:levels_left} and~\eqref{eq:levels_right} lead to inherent
differences in the left and right daughter nodes when splitting a mother node on
a categorical predictor in CART---differences that can have significant
ramifications when making predictions and inferences for observations with
absent levels.

\subsection{Random Forests}
\label{sec:rf}

Introduced in \citet{breiman:2001}, random forests are an ensemble learning
method that corrects for each individual tree's propensity to overfit the
training set.  This is accomplished through the use of bagging and a CART-like
tree learning algorithm in order to build a large collection of
``de-correlated'' decision trees.

\subsubsection{Bagging}

Proposed in \citet{breiman:1996a}, bagging is an ensembling technique for
improving the accuracy and stability of models.  Specifically, given a training
set
\[
  Z = \left\{(x_{1}, y_{1}), (x_{2}, y_{2}), \ldots, (x_{N}, y_{N})\right\},
\]
this is achieved by repeatedly sampling $N'$ observations with replacement from
$Z$ in order to generate $B$ bootstrapped training sets $Z_{1}, Z_{2}, \ldots,
Z_{B}$, where usually $N' = N$.  A separate model is then trained on each
bootstrapped training set $Z_{b}$, where we denote model $b$'s prediction on an
observation $x$ as $\hat{f}_{b}(x)$.  Here, showing each model a different
bootstrapped sample helps to de-correlate them, and the overall bagged estimate
$\hat{f}(x)$ for an observation $x$ can then be obtained by averaging over all
of the individual predictions in the case of regression
\[
\hat{f}(x) = \frac{1}{B}\sum_{b = 1}^{B}{\hat{f}_{b}(x)},
\]
or by taking the majority vote in the case of classification
\[
\hat{f}(x) = \argmax_{k \in \mathcal{K}}\!\left({\sum_{b = 1}^{B}{I\!\left(\hat{f}_{b}(x) = k\right)}}\right).
\]

One important aspect of bagging is the fact that each training observation $n$
will only appear ``in-bag'' in a subset of the bootstrapped training sets
$Z_{b}$.  Therefore, for each training observation $n$, an ``out-of-bag'' (OOB)
prediction can be constructed by only considering the subset of models in which
$n$ did not appear in the bootstrapped training set.  Moreover, an OOB error for
a bagged model can be obtained by evaluating the OOB predictions for all $N$
training observations---a performance metric which helps to alleviate the need
for cross-validation or a separate test set \citep{breiman:1996b}.

\subsubsection{CART-Like Tree Learning Algorithm}

In the case of random forests, the model that is being trained on each
individual bootstrapped training set $Z_{b}$ is a decision tree which is grown
using the CART methodology, but with two key modifications.

First, as mentioned previously in Section~\ref{sec:background}, the trees that
are grown in random forests are generally not pruned \citep{breiman:2001}.  And
second, instead of considering all $P$ predictors at a split, only a randomly
selected subset of the $P$ predictors is allowed to be used---a restriction
which helps to de-correlate the trees by placing a constraint on how similarly
they can be grown.  This process, which is known as the random subspace method,
was developed in \citet{amit:geman:1997} and \citet{ho:1998}.

\section{The Absent Levels Problem}
\label{sec:absent}

In Section~\ref{sec:intro}, we defined the absent levels problem as the inherent
issue for decision tree based methods occurring whenever there is an
indeterminacy over how to handle an observation that has reached a categorical
split which was determined when the observation in question's level was absent
during training, and we described the three different ways in which the absent
levels problem can arise.  Then, in Section~\ref{sec:unorderd_split}, we
discussed how the levels of a categorical predictor $p$ which were present in
the mother node $\mathcal{N}_{\mathcal{M}}$ during training were used to
determine its optimal splitting criterion $\mathcal{S}_{p}^{*}$.  In this
section, we investigate the potential consequences of overlooking the absent
levels problem where, for a categorical predictor $p$ with $Q$ unordered levels
which are indexed by the set $\mathcal{Q} = \left\{1, 2, \ldots, Q\right\}$, we
now also further denote the subset of the levels of $p$ that were present or
absent in the mother node $\mathcal{N}_{\mathcal{M}}$ during training,
respectively, as follows:
\begin{equation}
\begin{aligned}
\mathcal{Q}_{\mathcal{P}} &= \left\{q \in \mathcal{Q} : \left|\left\{n\in \mathcal{N}_{\mathcal{M}} : x_{np} = q\right\}\right| > 0\right\}, \\
\mathcal{Q}_{\mathcal{A}} &= \left\{q \in \mathcal{Q} : \left|\left\{n\in \mathcal{N}_{\mathcal{M}} : x_{np} = q\right\}\right| = 0\right\}.
\end{aligned}
\label{eq:present_absent}
\end{equation}
Specifically, by documenting how absent levels have been handled by Breiman and
Cutler's random forests \texttt{FORTRAN} code and the \texttt{randomForest}
\texttt{R} package, we show how failing to account for the absent levels problem
can systematically bias a model in practice.  However, although our
investigations are motivated by these two particular software implementations of
random forests, we emphasize that the absent levels problem is, first and
foremost, an intrinsic methodological issue for decision tree based methods.

\subsection{Regression}
\label{sec:rf_reg}

For regression trees using a squared error node impurity measure, recall from
our discussions in Section~\ref{sec:unorderd_split} and
equations~\eqref{eq:level_means}, \eqref{eq:levels_left}, and
\eqref{eq:levels_right}, that the split of a mother node
$\mathcal{N}_{\mathcal{M}}$ on a categorical predictor $p$ can be characterized
in terms of the splitting criterion's associated optimal numeric pseudo split
point $\tilde{s}_{p}^{*}$ and the means $\gamma_{p}(q)$ of the unordered levels
$q \in \mathcal{Q}$ of $p$ as follows:
\begin{itemize}
\item The unordered levels of $p$ being sent \textit{left} have means
  $\gamma_{p}(q)$ that are \textit{less than or equal to} $\tilde{s}_{p}^{*}$.
\item The unordered levels of $p$ being sent \textit{right} have means
  $\gamma_{p}(q)$ that are \textit{greater than} $\tilde{s}_{p}^{*}$.
\end{itemize}
Furthermore, recall from \eqref{eq:reg_pred}, that a node's prediction is given
by the mean of the training responses that are in the node.  Therefore, because
the prediction of each daughter node can be expressed as a weighted average over
the means $\gamma_{p}(q)$ of the present levels $q \in
\mathcal{Q}_{\mathcal{P}}$ that are being sent to it, it follows that
\textit{the left daughter node $\mathcal{N}_{\mathcal{L}}$ will always give a
  prediction that is smaller than the right daughter node
  $\mathcal{N}_{\mathcal{R}}$ when splitting on a categorical predictor $p$ in a
  regression tree that uses a squared error node impurity measure.}

In terms of execution, both the random forests \texttt{FORTRAN} code and the
\texttt{randomForest} {R} package employ the pseudo value procedure for
regression that was described in Section~\ref{sec:unorderd_split} when
determining the optimal splitting criterion $\mathcal{S}_{p}^{*}$ for a
categorical predictor $p$.  However, the code that is responsible for
calculating the mean $\gamma_{p}(q)$ within each unordered level $q \in
\mathcal{Q}$ as in equation~\eqref{eq:level_means} behaves as follows:
\[
\gamma_{p}(q) = \begin{cases}
  \text{ave}\!\left(y_{n} \mid n \in \mathcal{N}_{\mathcal{M}} \text{ and } x_{np} = q\right) &\text{if $q \in \mathcal{Q}_{\mathcal{P}}$} \\
0 &\text{if $q \in \mathcal{Q}_{\mathcal{A}}$}
\end{cases},
\]
where $\mathcal{Q}_{\mathcal{P}}$ and $\mathcal{Q}_{\mathcal{A}}$ are,
respectively, the present and absent levels of $p$ as defined in
\eqref{eq:present_absent}.

Although this ``zero imputation'' of the means $\gamma_{p}(q)$ for the absent
levels $q \in \mathcal{Q}_{\mathcal{A}}$ is inconsequential when determining the
optimal numeric pseudo split point $\tilde{s}_{p}^{*}$ during training, it can
be highly influential on the subsequent predictions that are made for
observations with absent levels.  In particular, by \eqref{eq:levels_left} and
\eqref{eq:levels_right}, the absent levels $q \in \mathcal{Q}_{\mathcal{A}}$
will be sent left if $\tilde{s}_{p}^{*} \geq 0$, and they will be sent right if
$\tilde{s}_{p}^{*} < 0$.  But, due to the systematic differences that exist
amongst the two daughter nodes, this arbitrary decision of sending the absent
levels left versus right can significantly impact the predictions that are made
on observations with absent levels---even though the model's final predictions
will also depend on any ensuing splits which take place after the absent levels
problem occurs, observations with absent levels will tend to be biased towards
smaller predictions when they are sent to the left daughter node, and they will
tend to be biased towards larger predictions when they are sent to the right
daughter node.

In addition, this behavior also implies that the random forest regression models
which are trained using either the random forests \texttt{FORTRAN} code or the
\texttt{randomForest} \texttt{R} package are sensitive to the set of possible
values that the training responses can take.  To illustrate, consider the
following two extreme cases when splitting a mother node
$\mathcal{N}_{\mathcal{M}}$ on a categorical predictor $p$:
\begin{itemize}
\item If the training responses $y_{n} > 0$ for all $n$, then the pseudo numeric
  split point $\tilde{s}_{p}^{*} > 0$ since the means $\gamma_{p}(q) > 0$ for
  all of the present levels $q \in \mathcal{Q}_{\mathcal{P}}$.  And because the
  ``imputed'' means $\gamma_{p}(q) = 0 < \tilde{s}_{p}^{*}$ for all $q \in
  \mathcal{Q}_{\mathcal{A}}$, the absent levels will always be sent to the left
  daughter node $\mathcal{N}_{\mathcal{L}}$ which gives smaller predictions.
\item If the training responses $y_{n} < 0$ for all $n$, then the pseudo numeric
  split point $\tilde{s}_{p}^{*} < 0$ since the means $\gamma_{p}(q) < 0$ for
  all of the present levels $q \in \mathcal{Q}_{\mathcal{P}}$.  And because the
  ``imputed'' means $\gamma_{p}(q) = 0 > \tilde{s}_{p}^{*}$ for all $q \in
  \mathcal{Q}_{\mathcal{A}}$, the absent levels will always be sent to the right
  daughter node $\mathcal{N}_{\mathcal{R}}$ which gives larger predictions.
\end{itemize}
And although this sensitivity to the training response values was most easily
demonstrated through these two extreme situations, the reader should not let
this overshadow the fact that the absent levels problem can also heavily
influence a model's performance in more general circumstances (e.g., when the
training responses are of mixed signs).

\subsection{Classification}
\label{sec:rf_class}

For binary classification trees using a weighted Gini index node impurity
measure, recall from our discussions in Section~\ref{sec:unorderd_split} and
equations~\eqref{eq:level_props}, \eqref{eq:levels_left}, and
\eqref{eq:levels_right}, that the split of a mother node
$\mathcal{N}_{\mathcal{M}}$ on a categorical predictor $p$ can be characterized
in terms of the splitting criterion's associated optimal numeric pseudo split
point $\tilde{s}_{p}^{*}$ and the $k = 1$ response class proportions
$\gamma_{p}(q)$ of the unordered levels $q \in \mathcal{Q}$ of $p$ as follows:
\begin{itemize}
\item The unordered levels of $p$ being sent \textit{left} have $k = 1$ response
  class proportions $\gamma_{p}(q)$ that are \textit{less than or equal to}
  $\tilde{s}_{p}^{*}$.
\item The unordered levels of $p$ being sent \textit{right} have $k = 1$
  response class proportions $\gamma_{p}(q)$ that are \textit{greater than}
  $\tilde{s}_{p}^{*}$.
\end{itemize}
In addition, recall from \eqref{eq:class_pred}, that a node's classification is
given by the response class that occurs the most amongst the training
observations that are in the node.  Therefore, because the response class
proportions of each daughter node can be expressed as a weighted average over
the response class proportions of the present levels $q \in
\mathcal{Q}_{\mathcal{P}}$ that are being sent to it, it follows that
\textit{the left daughter node $\mathcal{N}_{\mathcal{L}}$ is always less likely
  to classify an observation to the $\mathit{k = 1}$ response class than the
  right daughter node $\mathcal{N}_{\mathcal{R}}$ when splitting on a
  categorical predictor $p$ in a binary classification tree that uses a weighted
  Gini index node impurity measure.}

In terms of implementation, the \texttt{randomForest} \texttt{R} package uses
the pseudo value procedure for binary classification that was described in
Section~\ref{sec:unorderd_split} when determining the optimal splitting
criterion $\mathcal{S}_{p}^{*}$ for a categorical predictor $p$ with a ``large''
number of unordered levels.\footnote{The exact condition for using the pseudo
  value procedure for binary classification in version 4.6-12 of the
  \texttt{randomForest} \texttt{R} package is when a categorical predictor $p$
  has $Q > 10$ unordered levels.  Meanwhile, although the random forests
  \texttt{FORTRAN} code for binary classification references the pseudo value
  procedure, it does not appear to be implemented in the code.}  However, the
code that is responsible for computing the $k = 1$ response class proportion
$\gamma_{p}(q)$ within each unordered level $q \in \mathcal{Q}$ as in
equation~\eqref{eq:level_props} executes as follows:
\[
\gamma_{p}(q) = \begin{cases}
\frac{\left|\left\{n \,\in\, \mathcal{N}_{\mathcal{M}} \,:\, x_{np} \,=\, q \text{ and } y_{n} \,=\, 1\right\}\right|}{\left|\left\{n \,\in\, \mathcal{N}_{\mathcal{M}} \,:\, x_{np} \,=\, q\right\}\right|} &\text{if $q \in \mathcal{Q}_{\mathcal{P}}$} \\
0 &\text{if $q \in \mathcal{Q}_{\mathcal{A}}$}
\end{cases}.
\]
Therefore, the issues that arise here are similar to the ones that were
described for regression.

Even though this ``zero imputation'' of the $k = 1$ response class proportions
$\gamma_{p}(q)$ for the absent levels $q \in \mathcal{Q}_{\mathcal{A}}$ is
unimportant when determining the optimal numeric pseudo split point
$\tilde{s}_{p}^{*}$ during training, it can have a large effect on the
subsequent classifications that are made for observations with absent levels.
In particular, since the proportions $\gamma_{p}(q) \geq 0$ for all of the
present levels $q \in \mathcal{Q}_{\mathcal{P}}$, it follows from our
discussions in Section~\ref{sec:unorderd_split} that the numeric pseudo split
point $\tilde{s}_{p}^{*} \geq 0$.  And because the ``imputed'' proportions
$\gamma_{p}(q) = 0 \leq \tilde{s}_{p}^{*}$ for all $q \in
\mathcal{Q}_{\mathcal{A}}$, the absent levels will always be sent to the left
daughter node.  But, due to the innate differences that exist amongst the two
daughter nodes, this arbitrary choice of sending the absent levels left can
significantly affect the classifications that are made on observations with
absent levels---although the model's final classifications will also depend on
any successive splits which take place after the absent levels problem occurs,
the classifications for observations with absent levels will tend to be biased
towards the $k = 2$ response class.  Moreover, this behavior also implies that
the random forest binary classification models which are trained using the
\texttt{randomForest} \texttt{R} package may be sensitive to the actual ordering
of the response classes: since observations with absent levels are always sent
to the left daughter node $\mathcal{N}_{\mathcal{L}}$ which is more likely to
classify them to the $k = 2$ response class than the right daughter node
$\mathcal{N}_{\mathcal{R}}$, the classifications for these observations can be
influenced by interchanging the indices of the two response classes.

Meanwhile, for cases where the pseudo value procedure is not or cannot be used,
the random forests \texttt{FORTRAN} code and the \texttt{randomForest}
\texttt{R} package will instead adopt a more brute force approach that either
exhaustively or randomly searches through the space of possible splits.
However, to understand the potential problems that absent levels can cause in
these situations, we must first briefly digress into a discussion of how
categorical splits are internally represented in their code.

Specifically, in their code, a split on a categorical predictor $p$ is both
encoded and decoded as an integer whose binary representation identifies which
unordered levels go left (the bits that are ``turned on'') and which unordered
levels go right (the bits that are ``turned off'').  To illustrate, consider the
situation where a categorical predictor $p$ has four unordered levels, and where
the integer encoding of the split is $5$.  In this case, since $0101$ is the
binary representation of the integer $5$ (because $5 = [0] \cdot 2^{3} + [1]
\cdot 2^{2} + [0] \cdot 2^{1} + [1] \cdot 2^{0}$), levels $1$ and $3$ get sent
left while levels $2$ and $4$ get sent right.

Now, when executing an exhaustive search to find the optimal splitting criterion
$\mathcal{S}_{p}^{*}$ for a categorical predictor $p$ with $Q$ unordered levels,
the random forests \texttt{FORTRAN} code and the \texttt{randomForest}
\texttt{R} package will both follow the same systematic procedure:\footnote{The
  random forests \texttt{FORTRAN} code will use an exhaustive search for both
  binary and multiclass classification whenever $Q < 25$.  In version 4.6-12 of
  the \texttt{randomForest} \texttt{R} package, an exhaustive search will be
  used for both binary and multiclass classification whenever $Q < 10$.}
\textit{all $\mathit{2^{Q - 1} - 1}$ possible integer encodings for the
  non-redundant partitions of the unordered levels of predictor $\mathit{p}$ are
  evaluated in increasing sequential order starting from $\mathit{1}$ and ending
  at $\mathit{2^{Q -1} - 1}$, with the choice of the optimal splitting criterion
  $\mathit{\mathcal{S}_{p}^{*}}$ being updated if and only if the resulting
  weighted Gini index node impurity measure strictly improves}.

But since the absent levels $q \in \mathcal{Q}_{\mathcal{A}}$ are not present in
the mother node $\mathcal{N}_{\mathcal{M}}$ during training, \textit{sending
  them left or right has no effect on the resulting weighted Gini index}.  And
because turning on the bit for any particular level $q$ while holding the bits
for all of the other levels constant will always result in a larger integer, it
follows that \textit{the exhaustive search that is used by these two software
  implementations will always prefer splits that send all of the absent levels
  right since they are always checked before any of their analogous Gini index
  equivalent splits that send some of the absent levels left.}

Furthermore, in their exhaustive search, the leftmost bit corresponding to the
$Q^{th}$ indexed unordered level of a categorical predictor $p$ is always turned
off since checking the splits where this bit is turned on would be
redundant---they would amount to just swapping the ``left'' and ``right''
daughter node labels for splits that have already been evaluated.  Consequently,
the $Q^{th}$ indexed level of $p$ will also always be sent to the right daughter
node and, as a result, the classifications for observations with absent levels
will tend to be biased towards the response class distribution of the training
observations in the mother node $\mathcal{N}_{\mathcal{M}}$ that belong to this
$Q^{th}$ indexed level.  Therefore, although it may sound contradictory, this
also implies that the random forest multiclass classification models which are
trained using either the random forests \texttt{FORTRAN} code or the
\texttt{randomForest} \texttt{R} package may be sensitive to the actual ordering
of a categorical predictor's unordered levels---a reordering of these levels
could potentially interchange the ``left'' and ``right'' daughter node labels,
which could then subsequently affect the classifications that are made for
observations with absent levels since they will always be sent to whichever node
ends up being designated as the ``right'' daughter node.

Finally, when a categorical predictor $p$ has too many levels for an exhaustive
search to be computationally efficient, both the random forests \texttt{FORTRAN}
code and the \texttt{randomForest} \texttt{R} package will resort to
approximating the optimal splitting criterion $\mathcal{S}_{p}^{*}$ with the
best split that was found amongst a large number of randomly generated
splits.\footnote{The random forests \texttt{FORTRAN} code will use a random
  search for both binary and multiclass classification whenever $Q \geq 25$.  In
  version 4.6-12 of the \texttt{randomForest} \texttt{R} package, a random
  search will only be used when $Q \geq 10$ in the multiclass classification
  case.}  This is accomplished by randomly setting all of the bits in the binary
representations of the splits to either a 0 or a 1---a procedure which
ultimately results in each absent level being randomly sent to either the left
or right daughter node with equal probability.  As a result, although the absent
levels problem can still occur in these situations, it is difficult to determine
whether it results in any systematic bias.  However, it is still an open
question as to whether or not such a treatment of absent levels is sufficient.

\section{Heuristics for Mitigating the Absent Levels Problem}
\label{sec:heuristics}

Although a comprehensive theoretical analysis of the absent levels problem is
beyond the scope of this paper, in this section we briefly consider several
heuristics which may be able to help mitigate the issue.  Later, in
Section~\ref{sec:examples}, we empirically evaluate and compare how some of
these heuristics perform in practice when they are applied to three real data
examples.

\subsection{Missing Data Heuristics}

Even though absent levels are fully observed and known, the missing data
literature for decision tree based methods is still perhaps the area of existing
research that is most closely related to the absent levels problem.

\subsubsection{Stop}
One straightforward missing data strategy for dealing with absent levels would
be to simply stop an observation from going further down the tree whenever the
issue occurs and just use the mother node for prediction---a missing data
approach which has been adopted by both the \texttt{rpart} \texttt{R} package
for CART \citep{therneau:etal:2015} and the \texttt{gbm} \texttt{R} package for
generalized boosted regression models \citep{ridgeway:2013}.  Even with this
missing data functionality already in place, however, the \texttt{gbm}
\texttt{R} package has still had its own issues in readily extending it to the
case of absent levels---serving as another example of a software implementation
of a decision tree based method that has overlooked and suffered from the absent
levels problem.\footnote{See, for example,
  \url{https://code.google.com/archive/p/gradientboostedmodels/issues/7}}

\subsubsection{Distribution-Based Imputation (DBI)}
Another potential missing data technique would be to send an observation with an
absent level down both daughter nodes---perhaps by using the distribution-based
imputation (DBI) technique which is employed by the C4.5 algorithm for growing
decision trees \citep{quinlan:1993}.  In particular, an observation that
encounters an infeasible node split is first split into multiple
pseudo-instances, where each instance takes on a different imputed value and
weight based on the distribution of observed values for the splitting variable
in the mother node's subset of the training data.  These pseudo-instances are
then sent down their appropriate daughter nodes in order to proceed down the
tree as usual, and the final prediction is derived from the weighted predictions
of all the terminal nodes that are subsequently reached
\citep{saar-tsechansky:provost:2007}.

\subsubsection{Surrogate Splits}
Surrogate splitting, which the \texttt{rpart} \texttt{R} package also supports,
is arguably the most popular method of handling missing data in CART, and it may
provide another workable approach for mitigating the effects of absent levels.
Specifically, if $\left(p^{*}, \mathcal{S}_{p^{*}}^{*}\!\right)$ is found to be
the optimal splitting variable and splitting criterion pair for a mother node
$\mathcal{N}_{\mathcal{M}}$, then the first surrogate split is the $(p',
\mathcal{S}_{p'})$ pair where $p' \neq p^{*}$ that yields the split which most
closely mimics the optimal split's binary partitioning of
$\mathcal{N}_{\mathcal{M}}$, the second surrogate split is the $(p'',
\mathcal{S}_{p''})$ pair where $p'' \not\in \left\{p^{*}, p'\right\}$ resulting
in the second most similar binary partitioning of $\mathcal{N}_{\mathcal{M}}$ as
the optimal split, and so on.  Afterwards, when an observation reaches an
indeterminate split, the surrogates are tried in the order of decreasing
similarity until one of them becomes feasible \citep{breiman:etal:1984}.

However, despite its extensive use in CART, surrogate splitting may not be
entirely appropriate for ensemble tree methods like random forests.  As pointed
out in \citet{ishwaran:etal:2008}:
\begin{displayquote}
Although surrogate splitting works well for trees, the method may not be well
suited for forests.  Speed is one issue.  Finding a surrogate split is
computationally intensive and may become infeasible when growing a large number
of trees, especially for fully saturated trees used by forests.  Further,
surrogate splits may not even be meaningful in a forest paradigm.  [Random
  forests] randomly selects variables when splitting a node and, as such,
variables within a node may be uncorrelated, and a reasonable surrogate split
may not exist.  Another concern is that surrogate splitting alters the
interpretation of a variable, which affects measures such as [variable
  importance].
\end{displayquote}
Nevertheless, surrogate splitting is still available as a non-default option for
handling missing data in the \texttt{partykit} \texttt{R} package
\citep{hothorn:zeileis:2015}, which is an implementation of a bagging ensemble
of conditional inference trees that correct for the biased variable selection
issues which exist in several tree learning algorithms like CART and C4.5
\citep{hothorn:etal:2006}.

\subsubsection{Random/Majority}
The \texttt{partykit} \texttt{R} package also provides some other functionality
for dealing with missing data that may be applicable to the absent levels
problem.  These include the package's default approach of randomly sending the
observations to one of the two daughter nodes with the weighting done by the
number of training observations in each node or, alternatively, by simply having
the observations go to the daughter node with more training observations.
Interestingly, the \texttt{partykit} \texttt{R} package does appear to recognize
the possibility of absent levels occurring, and chooses to handle them as if
they were missing---its reference manual states that ``Factors in test samples
whose levels were empty in the learning sample are treated as missing when
computing predictions.''  Whether or not such missing data heuristics adequately
address the absent levels problem, however, is still unknown.

\subsection{Feature Engineering Heuristics}
Apart from missing data methods, feature engineering techniques which transform
the categorical predictors may also be viable approaches to mitigating the
effects of absent levels.

However, feature engineering techniques are not without their own drawbacks.
First, transforming the categorical predictors may not always be feasible in
practice since the feature space may become computationally unmanageable.  And
even when transformations are possible, they may further exacerbate variable
selection issues---many popular tree learning algorithms such as CART and C4.5
are known to be biased in favor of splitting on ordered predictors and
categorical predictors with many unordered levels since they offer more
candidate splitting points to choose from \citep{hothorn:etal:2006}.  Moreover,
by recoding a categorical predictor's unordered levels into several different
predictors, we forfeit a decision tree based method's natural ability to
simultaneously consider all of the predictor's levels together at a single
split.  Thus, it is not clear whether feature engineering techniques are
preferable when using decision tree based methods.

Despite these potential shortcomings, transformations of the categorical
predictors is currently required by the \texttt{scikit-learn} \texttt{Python}
module's implementation of random forests \citep{scikit-learn}.  There have,
however, been some discussions about extending the module so that it can support
the native categorical split capabilities used by the random forests
\texttt{FORTRAN} code and the \texttt{randomForest} \texttt{R}
package.\footnote{See, for example,
  \url{https://github.com/scikit-learn/scikit-learn/pull/3346}} But, needless to
say, such efforts would also have the unfortunate consequence of introducing the
indeterminacy of the absent levels problem into another popular software
implementation of a decision tree based method.

\subsubsection{One-Hot Encoding}
Nevertheless, one-hot encoding is perhaps the most straightforward feature
engineering technique that could be applied to the absent levels problem---even
though some unordered levels may still be absent when determining a categorical
split during training, any uncertainty over where to subsequently send these
absent levels would be eliminated by recoding the levels of each categorical
predictor into separate dummy predictors.

\section{Examples}
\label{sec:examples}

Although the actual severity of the absent levels problem will depend on the
specific data set and task at hand, in this section we present three real data
examples which illustrate how the absent levels problem can dramatically alter
the performance of decision tree based methods in practice.  In particular, we
empirically evaluate and compare how the seven different heuristics in the set
\[
\mathcal{H} =
\left\{\textit{Left}, \textit{Right}, \textit{Stop}, \textit{Majority}, \textit{Random},
\textit{DBI}, \textit{One-Hot} \right\}
\]
perform when confronted with the absent levels problem in random forests.

In particular, the first two heuristics that we consider in our set
$\mathcal{H}$ are the systematically biased approaches discussed in
Section~\ref{sec:absent} which have been employed by both the random forests
\texttt{FORTRAN} code and the \texttt{randomForest} \texttt{R} package due to
having overlooked the absent levels problem:
\begin{itemize}
\item \textsc{Left}: Sending the observation to the left daughter node.
\item \textsc{Right}: Sending the observation to the right daughter node.
\end{itemize}
Consequently, these two ``naive heuristics'' have been included in our analysis
for comparative purposes only.

In our set of heuristics $\mathcal{H}$, we also consider some of the missing
data strategies for decision tree based methods that we discussed in
Section~\ref{sec:heuristics}:
\begin{itemize}
\item \textsc{Stop}: Stopping the observation from going further down the
  tree and using the mother node for prediction.
\item \textsc{Majority}: Sending the observation to the daughter node with more
  training observations, with any ties being broken randomly.
\item \textsc{Random}: Randomly sending the observation to one of the two
  daughter nodes, with the weighting done by the number of training observations
  in each node.\footnote{We also investigated an alternative ``unweighted''
    version of the Random heuristic which randomly sends observations with
    absent levels to either the left or right daughter node with equal
    probability (analogous to the random search procedure that was described at
    the end of Section~\ref{sec:rf_class}).  However, because this unweighted
    version was found to be generally inferior to the ``weighted'' version
    described in our analysis, we have omitted it from our discussions for
    expositional clarity and conciseness.}
\item \textsc{Distribution-Based Imputation (DBI)}: Sending the observation to
  both daughter nodes using the C4.5 tree learning algorithm's DBI approach.
\end{itemize}
Unlike the two naive heuristics, these ``missing data heuristics'' are all less
systematic in their preferences amongst the two daughter nodes.

Finally, in our set $\mathcal{H}$, we also consider a ``feature engineering
heuristic'' which transforms all of the categorical predictors in the original
data set:
\begin{itemize}
\item \textsc{One-Hot}: Recoding every categorical predictor's set of possible
  unordered levels into separate dummy predictors
\end{itemize}
Under this heuristic, although unordered levels may still be absent when
determining a categorical split during training, there is no longer any
uncertainty over where to subsequently send observations with absent levels.

Code for implementing the naive and missing data heuristics was built on top of
version 4.6-12 of the \texttt{randomForest} \texttt{R} package.  Specifically,
the \texttt{randomForest} \texttt{R} package is used to first train the random
forest models as usual.  Afterwards, each individual tree's in-bag training data
is sent back down the tree according to the tree's set of decision rules in
order to record the unordered levels that were absent at each categorical split.
Finally, when making predictions or inferences, our code provides some
functionality for carrying out each of the naive and missing data heuristics
whenever the absent levels problem occurs.

Each of the random forest models that we consider in our analysis is trained
``off-the-shelf'' by using the \texttt{randomForest} \texttt{R} package's
default settings for the algorithm's tuning parameters.  Moreover, to account
for the inherent randomness in the random forests algorithm, we repeat each of
our examples $1000$ times with a different random seed used to initialize each
experimental replication.  However, because of the way in which we have
structured our code, we note that our analysis is able to isolate the effects of
the naive and missing data heuristics on the absent levels problem since, within
each experimental replication, their underlying random forest models are
identical with respect to each tree's in-bag training data and differ only in
terms of their treatment of the absent levels.  As a result, the predictions and
inferences obtained from the naive and missing data heuristics will be
positively correlated across the $1000$ experimental replications that we
consider for each example---a fact which we exploit in order to improve the
precision of our comparisons.

\begin{figure}[t]
  \centering
  \includegraphics[scale=.395]{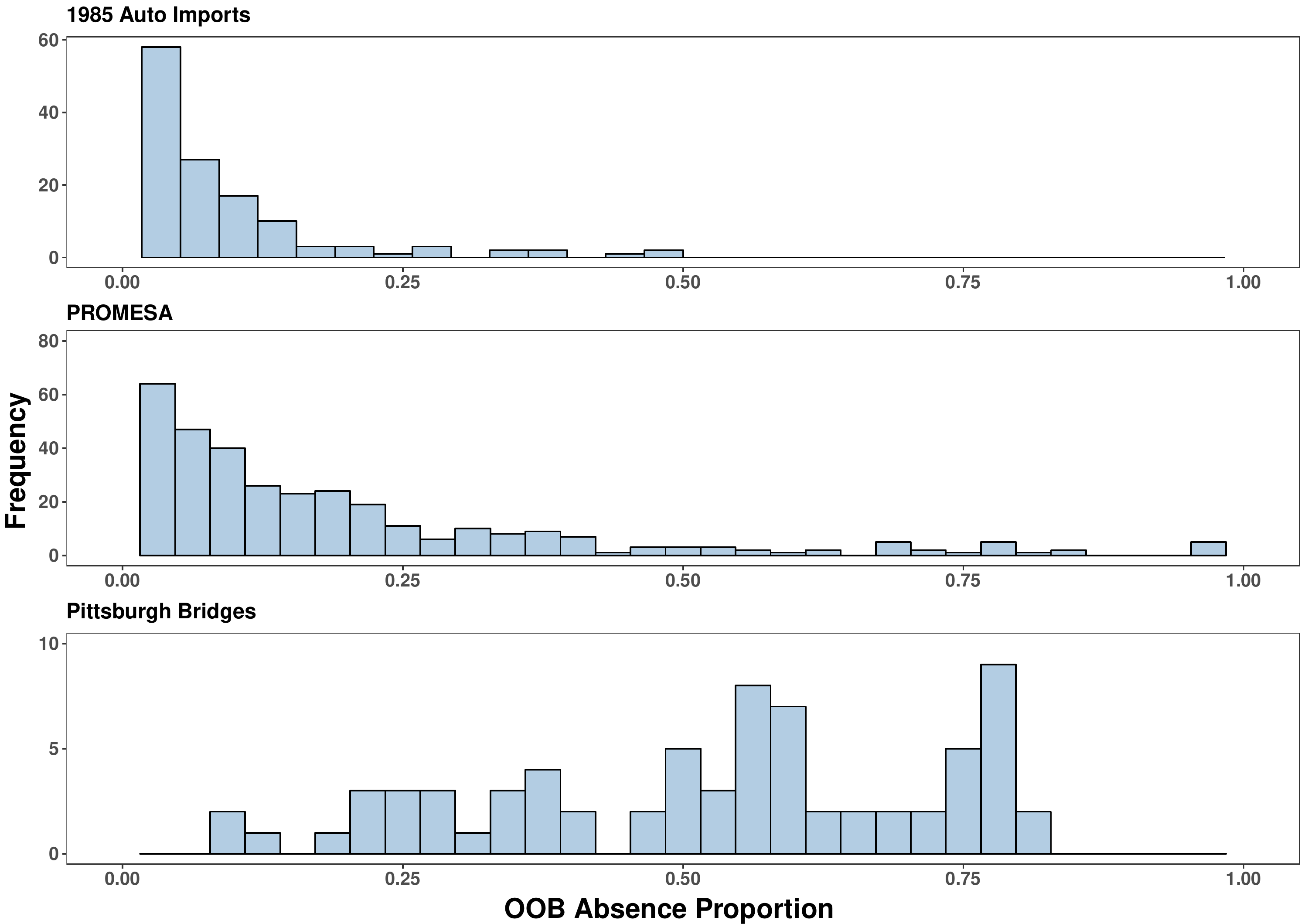}
  \captionsetup{format = hang}
  \caption{Histograms for each example's distribution of OOB absence
    proportions.}
  \label{fig:abs_props}
\end{figure}

Recall from Section~\ref{sec:heuristics}, however, that the random forest models
which are trained on feature engineered data sets are intrinsically different
from the random forest models which are trained on their original untransformed
data set counterparts.  Therefore, although we use the same default
\texttt{randomForest} \texttt{R} package settings and the same random seed to
initialize each of the One-Hot heuristic's experimental replications, we note
that its predictions and inferences will be essentially uncorrelated with the
naive and missing data heuristics across each example's $1000$ experimental
replications.

\subsection{1985 Auto Imports}

For a regression example, we consider the 1985 Auto Imports data set from the
UCI Machine Learning Repository \citep{lichman:2013} which, after discarding
observations with missing data, contains $25$ predictors that can be used to
predict the prices of $159$ cars.  Categorical predictors for which the absent
levels problem can occur include a car's make (18 levels), body style (5
levels), drive layout (3 levels), engine type (5 levels), and fuel system (6
levels).  Furthermore, because all of the car prices are positive, we know from
Section~\ref{sec:rf_reg} that the random forests \texttt{FORTRAN} code and the
\texttt{randomForest} \texttt{R} package will both always employ the Left
heuristic when faced with absent levels for this particular data
set.\footnote{This is the case for versions 4.6-7 and earlier of the
  \texttt{randomForest} \texttt{R} package.  Beginning in version 4.6-9,
  however, the \texttt{randomForest} \texttt{R} package began to internally mean
  center the training responses prior to fitting the model, with the mean being
  subsequently added back to the predictions of each node.  Consequently, the
  Left heuristic isn't always used in these versions of the
  \texttt{randomForest} \texttt{R} package since the training responses that the
  model actually considers are of mixed sign.  Nevertheless, such a strategy
  still fails to explicitly address the underlying absent levels problem.}

The top panel in Figure~\ref{fig:abs_props} depicts a histogram of this
example's OOB absence proportions, which we define for each training observation
as the proportion of its OOB trees across all $1000$ experimental replications
which had the absent levels problem occur at least once when using the training
set with the original untransformed categorical predictors.  Meanwhile,
Table~\ref{tab:abs_props} provides a more detailed summary of this example's
distribution of OOB absence proportions.  Consequently, although there is a
noticeable right skew in the distribution, we see that most of the observations
in this example had the absent levels problem occur in less than $5\%$ of their
OOB trees.

\begin{table}[t]
\begin{center}
\small
\begin{tabular}{l | c c c}
\specialrule{.1em}{.05em}{.05em}
\textbf{Statistic} & \textbf{1985 Auto Imports} & \textbf{PROMESA} & \textbf{Pittsburgh Bridges} \\
\hline
Min                        & 0.003 & 0.001 & 0.080        \\
1st Quartile               & 0.021 & 0.020 & 0.368        \\
Median                     & 0.043 & 0.088 & 0.564        \\
Mean                       & 0.076 & 0.162 & 0.526        \\
3rd Quartile               & 0.093 & 0.206 & 0.702        \\
Max                        & 0.498 & 0.992 & 0.820        \\
\specialrule{.1em}{.05em}{.05em}
\end{tabular}
\end{center}
\captionsetup{format = hang}
\caption{Summary statistics for each example's distribution of OOB absence
  proportions.}
\label{tab:abs_props}
\end{table}

\begin{sidewaysfigure}
  \centering
  \includegraphics[scale=.60]{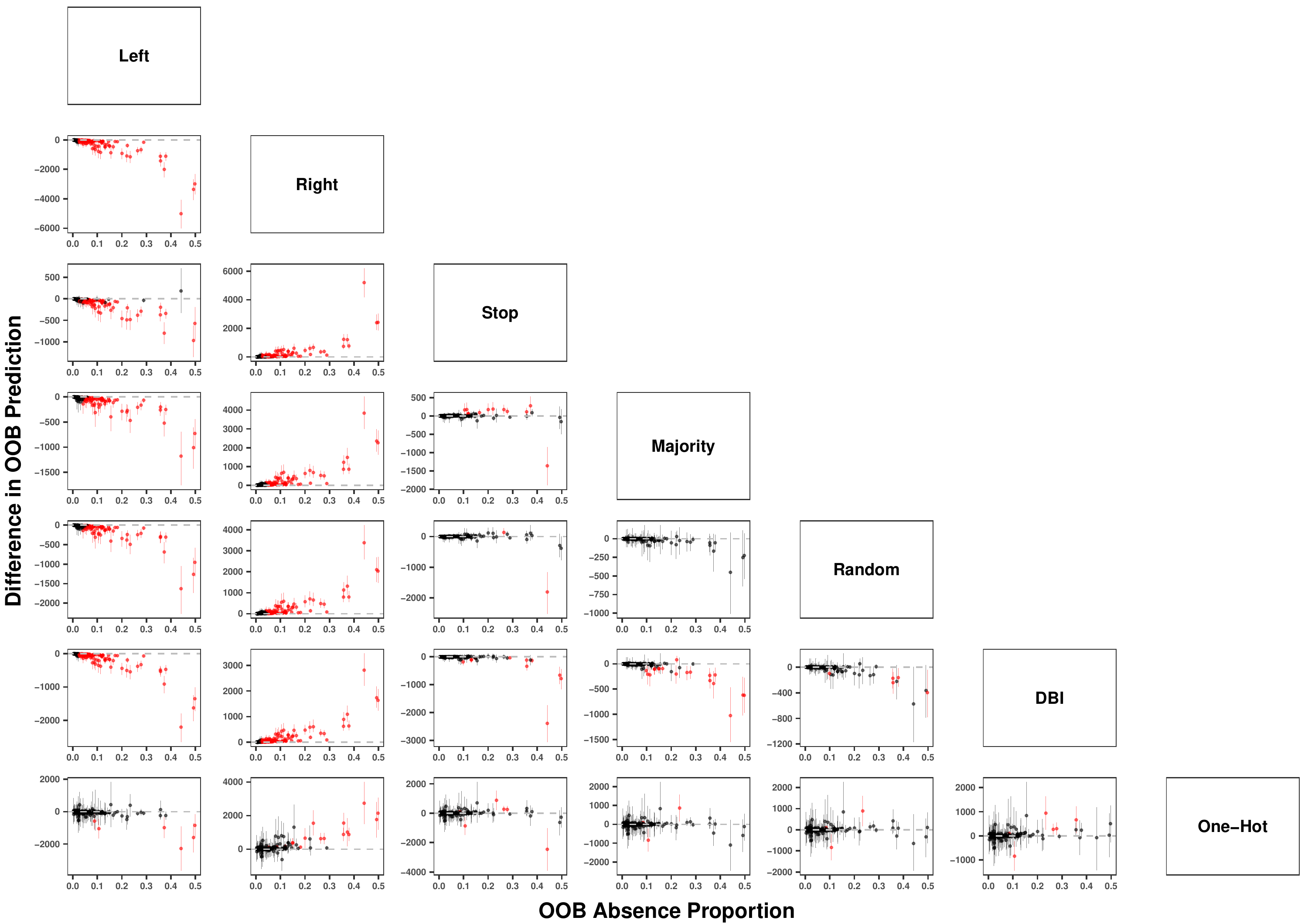}
  \captionsetup{format = hang}
  \caption{Pairwise differences in the OOB predictions as a function of the OOB
    absence proportions in the 1985 Auto Imports data set.  Each panel plots the
    mean and middle $95\%$ of the differences across all $1000$ experimental
    replications when the OOB predictions of the heuristic that is labeled at
    the right of the panel's row is subtracted from the OOB predictions of the
    heuristic that is labeled at the top of the panel's column.  Differences
    were taken within each experimental replication in order to account for the
    positive correlation that exists between the naive and missing data
    heuristics.  Intervals containing zero (the horizontal dashed line) are in
    black, while intervals not containing zero are in red.}
  \label{fig:imports85_diffs_plot}
\end{sidewaysfigure}

Let $\hat{y}_{nr}^{\left(h\right)}$ denote the OOB prediction that a heuristic
$h$ makes for an observation $n$ in an experimental replication $r$.  Then,
within each experimental replication $r$, we can compare the predictions that
two different heuristics $h_{1}, h_{2} \in \mathcal{H}$ make for an observation
$n$ by considering the difference $\hat{y}_{nr}^{\left(h_{1}\right)} -
\hat{y}_{nr}^{\left(h_{2}\right)}$.  We summarize these comparisons for all
possible pairwise combinations of the seven heuristics in
Figure~\ref{fig:imports85_diffs_plot}, where each panel plots the mean and
middle $95\%$ of these differences across all $1000$ experimental replications
as a function of the OOB absence proportion.  From the red intervals in
Figure~\ref{fig:imports85_diffs_plot}, we see that significant differences in
the predictions of the heuristics do exist, with the magnitude of the point
estimates and the width of the intervals tending to increase with the OOB
absence proportion---behavior that agrees with our intuition that the
distinctive effects of each heuristic should become more pronounced the more
often the absent levels problem occurs.

\begin{figure}[t]
  \centering
  \includegraphics[scale=.425]{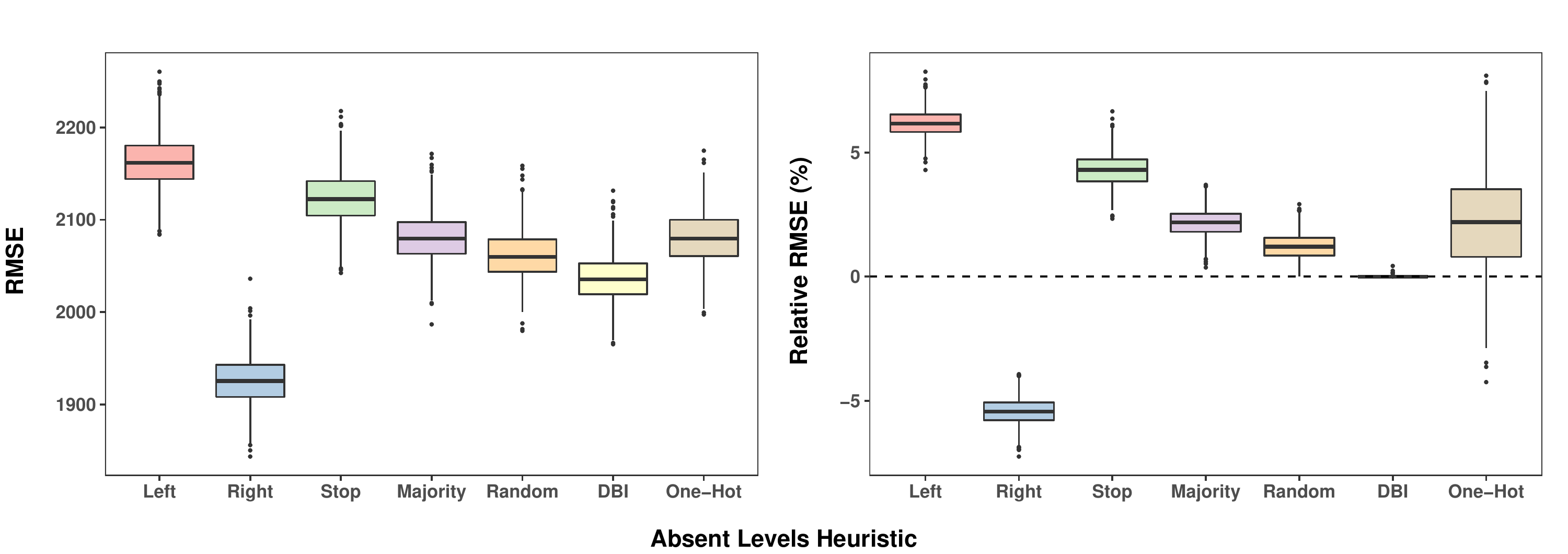}
  \captionsetup{format = hang}
  \caption{RMSEs for the OOB predictions of the seven heuristics in the 1985 Auto
    Imports data set.  The left panel shows boxplots of each heuristic's
    marginal distribution of RMSEs across all $1000$ experimental replications,
    which ignores the positive correlation that exists between the naive and
    missing data heuristics.  The right panel accounts for this positive
    correlation by comparing the RMSEs of the heuristics relative to the best
    RMSE that was obtained amongst the missing data heuristics within
    each of the $1000$ experimental replications as in \eqref{eq:rrmse}.}
  \label{fig:imports85_loss_plot}
\end{figure}

In addition, we can evaluate the overall performance of each heuristic $h \in
\mathcal{H}$ within an experimental replication $r$ in terms of its root mean
squared error (RMSE):
\[
\text{RMSE}_{r}^{\left(h\right)} = \sqrt{\frac{1}{N}\sum_{n = 1}^{N}{\left(y_{n} -
    \hat{y}_{nr}^{\left(h\right)}\right)^2}}.
\]
Boxplots displaying each heuristic's marginal distribution of RMSEs across all
$1000$ experimental replications are shown in the left panel of
Figure~\ref{fig:imports85_loss_plot}.  However, these marginal boxplots ignore
the positive correlation that exists between the naive and missing data
heuristics.  Therefore, within every experimental replication $r$, we also
compare the RMSE for each heuristic $h \in \mathcal{H}$ relative to the best
RMSE that was achieved amongst the missing data heuristics $\mathcal{H}_{m} =
\left\{\textit{Stop}, \textit{Majority}, \textit{Random},
\textit{DBI}\,\right\}$:
\begin{equation}
  \text{RMSE}_{r}^{\left(h \relative \mathcal{H}_{m}\right)} = \frac{\text{RMSE}_{r}^{\left(h\right)} - \min_{h \in \mathcal{H}_{m}}\text{RMSE}_{r}^{\left(h\right)}}{\min_{h \in \mathcal{H}_{m}}\text{RMSE}_{r}^{\left(h\right)}}.
  \label{eq:rrmse}
\end{equation}
Here we note that the Left and Right heuristics were not considered in the
definition of the best RMSE achieved within each experimental replication $r$
due to the issues discussed in Section~\ref{sec:absent}, while the One-Hot
heuristic was excluded from this definition since it is essentially uncorrelated
with the other six heuristics across all $1000$ experimental replications.
Boxplots of these relative RMSEs are shown in the right panel of
Figure~\ref{fig:imports85_loss_plot}.

\subsubsection{Naive Heuristics}

Relative to all of the other heuristics and consistent with our discussions in
Section~\ref{sec:rf_reg}, we see from Figure~\ref{fig:imports85_diffs_plot} that
the Left and Right heuristics have a tendency to severely underpredict and
overpredict, respectively.  Furthermore, for this particular example, we notice
from Figure~\ref{fig:imports85_loss_plot} that the random forests
\texttt{FORTRAN} code and the \texttt{randomForest} \texttt{R} package's
behavior of always sending absent levels left in this particular data set
substantially underperforms relative to the other heuristics---it gives an
\text{RMSE} that is, on average, $6.2\%$ worse than the best performing missing
data heuristic.  And although the Right heuristic appears to perform
exceptionally well, we again stress the misleading nature of this
performance---its tendency to overpredict just coincidentally happens to be
beneficial in this specific situation.

\subsubsection{Missing Data Heuristics}

As can be seen from Figure~\ref{fig:imports85_diffs_plot}, the predictions
obtained from the four missing data heuristics are more aligned with one another
than they are with the Left, Right, and One-Hot heuristics.  Considerable
disparities in their predictions do still exist, however, and from
Figure~\ref{fig:imports85_loss_plot} we note that amongst the four missing data
heuristics, the DBI heuristic clearly performs the best.  And although the
Majority heuristic fares slightly worse than the Random heuristic, they both
perform appreciably better than the Stop heuristic.

\subsubsection{Feature Engineering Heuristic}

Recall that the One-Hot heuristic is essentially uncorrelated with the other six
heuristics across all $1000$ experimental replications---a fact which is
reflected in its noticeably wider intervals in
Figure~\ref{fig:imports85_diffs_plot} and in its larger relative RMSE boxplot in
Figure~\ref{fig:imports85_loss_plot}.  Nevertheless, it can still be observed
from Figure~\ref{fig:imports85_loss_plot} that although the One-Hot heuristic's
predictions are sometimes able to outperform the other heuristics, on average,
it yields an RMSE that is $2.2\%$ worse than than the best performing missing
data heuristic.

\subsection{PROMESA}

For a binary classification example, we consider the June 9, 2016 United States
House of Representatives vote on the Puerto Rico Oversight, Management, and
Economic Stability Act (PROMESA) for addressing the Puerto Rican government's
debt crisis.  Data for this vote was obtained by using the \texttt{Rvoteview}
\texttt{R} package to query the Voteview database \citep{lewis:2015}.  After
omitting those who did not vote on the bill, the data set contains four
predictors that can be used to predict the binary ``No'' or ``Yes'' votes of
$424$ House of Representative members.  These predictors include a categorical
predictor for a representative's political party (2 levels), a categorical
predictor for a representative's state (50 levels), and two ordered predictors
which quantify aspects of a representative's political ideological position
\citep{mccarty:1997}.

The ``No'' vote was taken to be the $k = 1$ response class in our analysis,
while the ``Yes'' vote was taken to be the $k = 2$ response class. Recall from
Section~\ref{sec:rf_class}, that this ordering of the response classes is
meaningful in a binary classification context since the \texttt{randomForest}
\texttt{R} package will always use the Left heuristic which biases predictions
for observations with absent levels towards whichever response class is indexed
by $k = 2$ (corresponding to the ``Yes'' vote in our analysis).

\begin{sidewaysfigure}
  \centering
  \includegraphics[scale=.60]{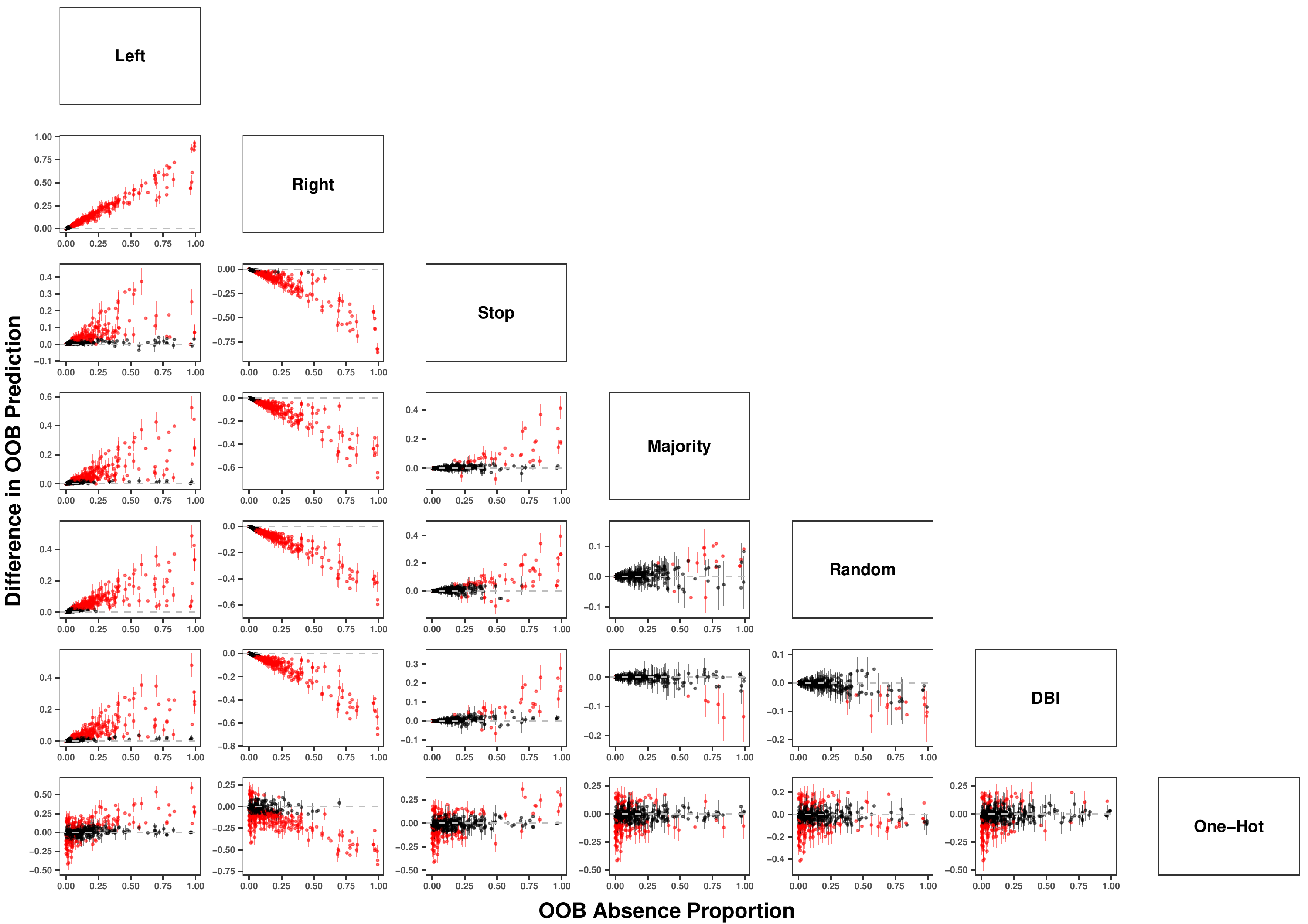}
  \captionsetup{format = hang}
  \caption{Pairwise differences in the OOB predicted probabilities of voting
    ``Yes'' as a function of the OOB absence proportion in the PROMESA data set.
    Each panel plots the mean and middle $95\%$ of the pairwise differences
    across all $1000$ experimental replications when the OOB predicted
    probabilities of the heuristic that is labeled at the right of the panel's
    row is subtracted from the OOB predicted probabilities of the heuristic that
    is labeled at the top of the panel's column.  Differences were taken within
    each experimental replication to account for the positive correlation that
    exists between the naive and missing data heuristics.  Intervals containing
    zero (the horizontal dashed line) are in black, while intervals not
    containing zero are in red.}
  \label{fig:puerto_diffs_plot}
\end{sidewaysfigure}

From Figure~\ref{fig:abs_props} and Table~\ref{tab:abs_props}, we see that the
absent levels problem occurs much more frequently in this example than it did in
our 1985 Auto Imports example.  In particular, the seven House of Representative
members who were the sole representatives from their state had OOB absence
proportions that were greater than $0.961$ since the absent levels problem
occurred for these observations every time they reached an OOB tree node that
was split on the state predictor.

For random forest classification models, the predicted probability that an
observation belongs to a response class $k$ can be estimated by the proportion
of the observation's trees which classify it to class $k$.\footnote{This is the
  approach that is used by the \texttt{randomForest} \texttt{R} package.  The
  \texttt{scikit-learn} \texttt{Python} module uses an alternative method of
  calculating the predicted response class probabilities which takes the average
  of the predicted class probabilities over the trees in the random forest,
  where the predicted probability of a response class $k$ in an individual tree
  is estimated using the proportion of a node's training samples that belong to
  the response class $k$.}  Let $\hat{p}_{nkr}^{\left(h\right)}$ denote the OOB
predicted probability that a heuristic $h$ assigns to an observation $n$ of
belonging to a response class $k$ in an experimental replication $r$.  Then,
within each experimental replication $r$, we can compare the predicted
probabilities that two different heuristics $h_{1}, h_{2} \in \mathcal{H}$
assign to an observation $n$ by considering the difference
$\hat{p}_{nkr}^{\left(h_{1}\right)} - \hat{p}_{nkr}^{\left(h_{2}\right)}$.  We
summarize these differences in the predicted probabilities of voting ``Yes'' for
all possible pairwise combinations of the seven heuristics in
Figure~\ref{fig:puerto_diffs_plot}, where each panel plots the mean and middle
$95\%$ of these differences across all $1000$ experimental replications as a
function of the OOB absence proportion.

\begin{sidewaysfigure}
  \centering
  \includegraphics[scale=.60]{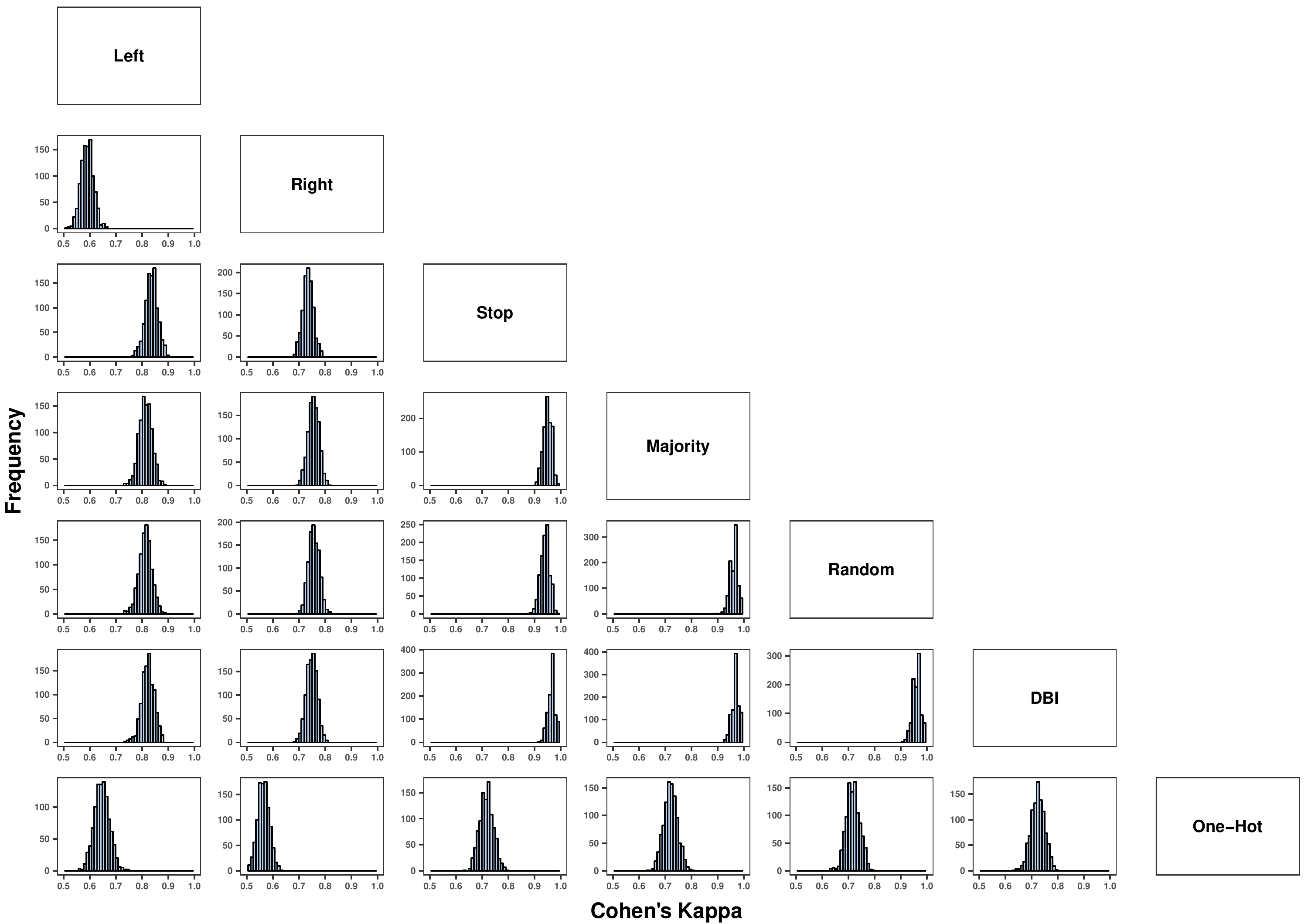}
  \captionsetup{format = hang}
  \caption{Pairwise Cohen's kappa coefficients for the seven different
    heuristics as defined in \eqref{eq:kappa} when the random forests
    algorithm's default majority vote discrimination threshold of $0.5$ is used
    in the PROMESA data set.  Each panel plots the histogram of coefficients
    across all $1000$ experimental replications when the OOB classifications of
    the heuristic that is labeled at the right of the panel's row are compared
    against the OOB classifications of the heuristic that is labeled at the top
    of the panel's column.  Cohen's kappa coefficients were calculated within
    each of the $1000$ experimental replications to account for the positive
    correlation between the naive and missing data heuristics.}
  \label{fig:puerto_kappas_plot}
\end{sidewaysfigure}

The large discrepancies in the predicted probabilities that are observed in
Figure~\ref{fig:puerto_diffs_plot} are particularly concerning since they can
lead to different classifications.  If we let $\hat{y}_{nr}^{\left(h\right)}$
denote the OOB classification that a heuristic $h$ makes for an observation $n$
in an experimental replication $r$, then Cohen's kappa coefficient
\citep{cohen:1960} provides one way of measuring the level of agreement between
two different heuristics $h_{1}, h_{2} \in \mathcal{H}$:
\begin{equation}
\kappa_{r}^{(h_{1}, h_{2})} = \frac{o_{r}^{(h_{1}, h_{2})} - e_{r}^{(h_{1}, h_{2})}}{1 - e_{r}^{(h_{1}, h_{2})}},
\label{eq:kappa}
\end{equation}
where
\[
o_{r}^{(h_{1}, h_{2})} = \frac{1}{N}\sum_{n = 1}^{N}{I\!\left(\hat{y}_{nr}^{(h_{1})} = \hat{y}_{nr}^{(h_{2})}\right)}
\]
is the observed probability of agreement between the two heuristics, and where
\[
e_{r}^{(h_{1}, h_{2})} = \frac{1}{N^2}\sum_{k = 1}^{K}{\left[\left(\sum_{n = 1}^{N}{I\!\left(\hat{y}_{nr}^{(h_{1})} = k\right)}\right) \cdot \left(\sum_{n = 1}^{N}{I\!\left(\hat{y}_{nr}^{(h_{2})} = k\right)}\right)\right]}
\]
is the expected probability of the two heuristics agreeing by chance.
Therefore, within an experimental replication $r$, we will observe
$\kappa_{r}^{(h_{1}, h_{2})} = 1$ if the two heuristics are in complete
agreement, and we will observe $\kappa_{r}^{(h_{1}, h_{2})} \approx 0$ if there
is no agreement amongst the two heuristics other than what would be expected by
chance.  In Figure~\ref{fig:puerto_kappas_plot}, we plot histograms of the
Cohen's kappa coefficient for all possible pairwise combinations of the seven
heuristics across all $1000$ experimental replications when the random forests
algorithm's default majority vote discrimination threshold of $0.5$ is used.

More generally, the areas underneath the receiver operating characteristic (ROC)
and precision-recall (PR) curves can be used to compare the overall performance
of binary classifiers as the discrimination threshold is varied between 0 and 1.
Specifically, as the discrimination threshold changes, the ROC curve plots the
proportion of positive observations that a classifier correctly labels as a
function of the proportion of negative observations that a classifier
incorrectly labels, while the PR curve plots the proportion of a classifier's
positive labels that are truly positive as a function of the proportion of
positive observations that a classifier correctly labels
\citep{davis:goadrich:2006}.

\begin{figure}[t]
  \centering
  \includegraphics[scale=.425]{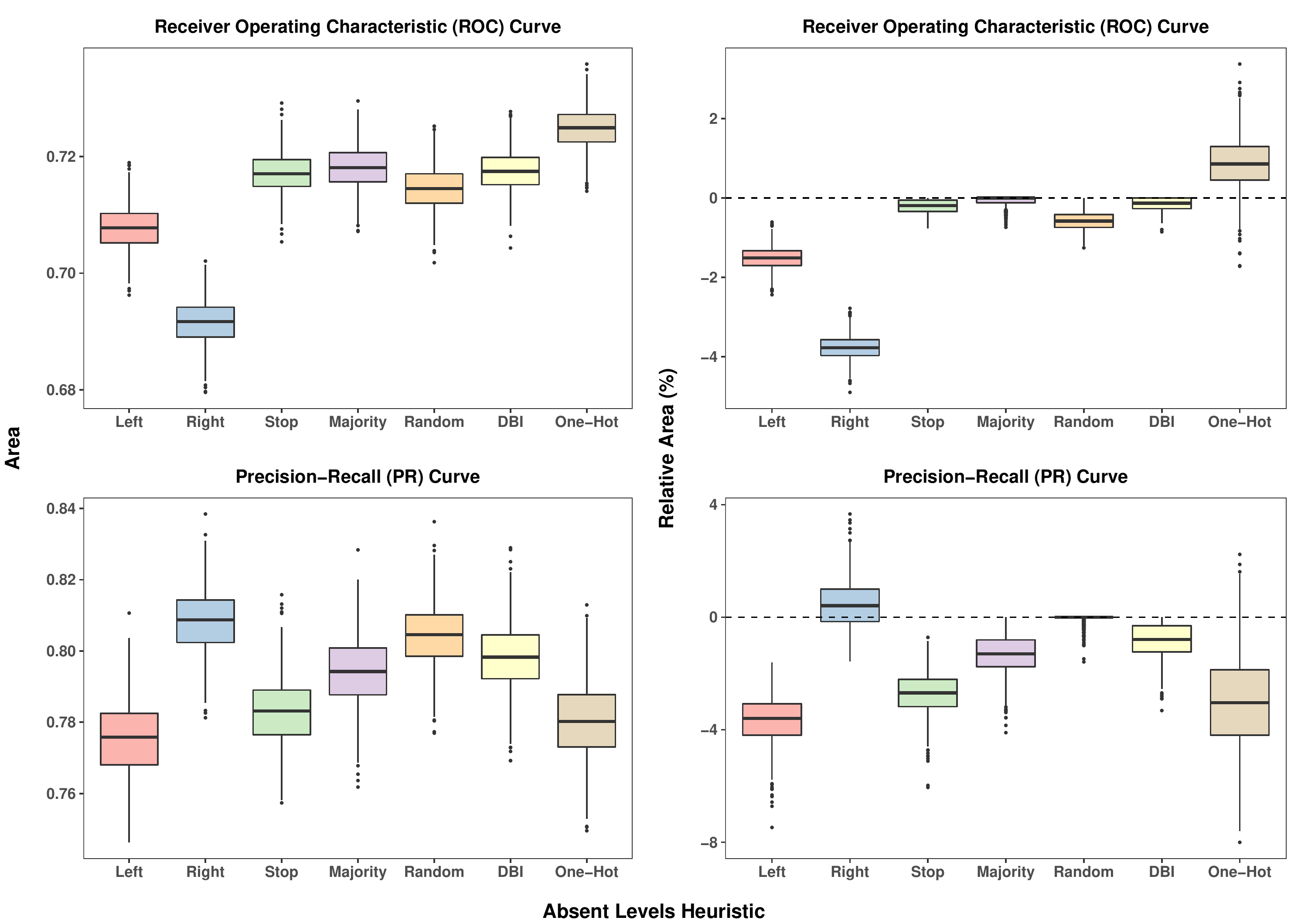}
  \captionsetup{format = hang}
  \caption{Areas underneath the ROC and PR curves for the seven heuristics in
    the PROMESA data set.  The left panels show boxplots of each heuristic's
    marginal distribution of areas across all $1000$ experimental replications,
    which ignores the positive correlation that exists between the naive and
    missing data heuristics.  The right panels account for this positive
    correlation by comparing the areas of the heuristics relative to the best
    area that was obtained amongst the missing data heuristics within each of
    the $1000$ experimental replications as in \eqref{eq:rauc}.}
  \label{fig:puerto_area_plot}
\end{figure}

Taking the ``Yes'' vote to be the positive response class in our analysis, we
calculate the areas underneath the ROC and PR curves for each heuristic $h \in
\mathcal{H}$ within each experimental replication $r$.  Boxplots depicting each
heuristic's marginal distribution of these two areas across all $1000$
experimental replications are shown in the left panels of
Figure~\ref{fig:puerto_area_plot}.  However, these marginal boxplots ignore the
positive correlation that exists between the naive and missing data heuristics.
Therefore, within every experimental replication $r$ and similar to what was
previously done in our 1985 Auto Imports example, we also compare the areas for
each heuristic $h \in \mathcal{H}$ relative to the best area that was achieved
amongst the missing data heuristics $\mathcal{H}_{m} = \left\{\textit{Stop},
\textit{Majority}, \textit{Random}, \textit{DBI}\,\right\}$:
\begin{equation}
  \text{AUC}_{r}^{\left(h \relative \mathcal{H}_{m}\right)} = \frac{\text{AUC}_{r}^{\left(h\right)} - \max_{h \in \mathcal{H}_{m}}\text{AUC}_{r}^{\left(h\right)}}{\max_{h \in \mathcal{H}_{m}}\text{AUC}_{r}^{\left(h\right)}},
\label{eq:rauc}
\end{equation}
where, depending on the context, $\text{AUC}_{r}^{(h)}$ denotes the area that is
underneath either the ROC or PR curve for heuristic $h$ in experimental
replication $r$.  Boxplots of these relative areas across all $1000$
experimental replications are displayed in the right panels of
Figure~\ref{fig:puerto_area_plot}.

\subsubsection{Naive Heuristics}

As expected given our discussions in Section~\ref{sec:rf_class} and how we have
chosen to index the response classes in our analysis, we see from
Figure~\ref{fig:puerto_diffs_plot} that the Left heuristic results in
significantly higher predicted probabilities of voting ``Yes'' than the other
heuristics, while the Right heuristic yields predicted probabilities of voting
``Yes'' that are substantially lower.  The consequences of this behavior in
terms of making classifications can be observed in
Figure~\ref{fig:puerto_kappas_plot}, where we note that both the Left and Right
heuristics tend to exhibit a high level of disagreement when compared against
any other heuristic's classifications.  Moreover,
Figure~\ref{fig:puerto_area_plot} illustrates that the \texttt{randomForest}
\texttt{R} package's practice of always sending absent levels left in binary
classification is noticeably detrimental here---relative to the best performing
missing data heuristic, it gives areas underneath the ROC and PR curves that
are, on average, $1.5\%$ and $3.7\%$ worse, respectively.  And although the
Right heuristic appears to do well in terms of the area underneath the PR curve,
we once again emphasize the spurious nature of this performance and caution
against taking it at face value.

\subsubsection{Missing Data Heuristics}

Similar to what was previously seen in our 1985 Auto Imports example,
Figures~\ref{fig:puerto_diffs_plot} and \ref{fig:puerto_kappas_plot} show that
the four missing data heuristics tend to exhibit a higher level of agreement
with one another than they do with the Left, Right, and One-Hot heuristics.
However, significant differences do still exist, and we see from
Figure~\ref{fig:puerto_area_plot} that the relative performances of the
heuristics will vary depending on the specific task at hand---the Majority
heuristic slightly outperforms the three other missing data heuristics in terms
of the area underneath the ROC curve, while the Random heuristic does
considerably better than all of its missing data counterparts with respect to
the area underneath the PR curve.

\subsubsection{Feature Engineering Heuristic}

Although it is essentially uncorrelated with the other six heuristics across all
$1000$ experimental replications, Figures~\ref{fig:puerto_diffs_plot}
and~\ref{fig:puerto_kappas_plot} still suggest that the One-Hot heuristic's
predicted probabilities and classifications can greatly differ from the other
six heuristics.  Furthermore, Figure~\ref{fig:puerto_area_plot} shows that even
though the One-Hot heuristic may appear to perform well in terms of the area
underneath its ROC curve relative to the missing data heuristics, its
performance in the PR context is rather lackluster.

\subsection{Pittsburgh Bridges}

For a multiclass classification example, we consider the Pittsburgh Bridges data
set from the UCI Machine Learning Repository which, after removing observations
with missing data, contains seven predictors that can be used to classify $72$
bridges to one of seven different bridge types.  The categorical predictors in
this data set for which the absent levels problem can occur include a bridge's
river (3 levels), purpose (3 levels), and location (46 levels).  Consequently,
recall from Section~\ref{sec:rf_class}, that the random forests \texttt{FORTRAN}
code and \texttt{randomForest} \texttt{R} package will both employ an exhaustive
search that always sends absent levels right when splitting on either the river
or purpose predictors, and that they will both resort to using a random search
that sends absent levels either left or right with equal probability when
splitting on the location predictor since it has too many levels for an
exhaustive search to be computationally efficient.  The OOB absence proportions
for this example are summarized in the bottom panel of
Figure~\ref{fig:abs_props} and in Table~\ref{tab:abs_props}.

\begin{figure}[t]
  \centering
  \includegraphics[scale=.425]{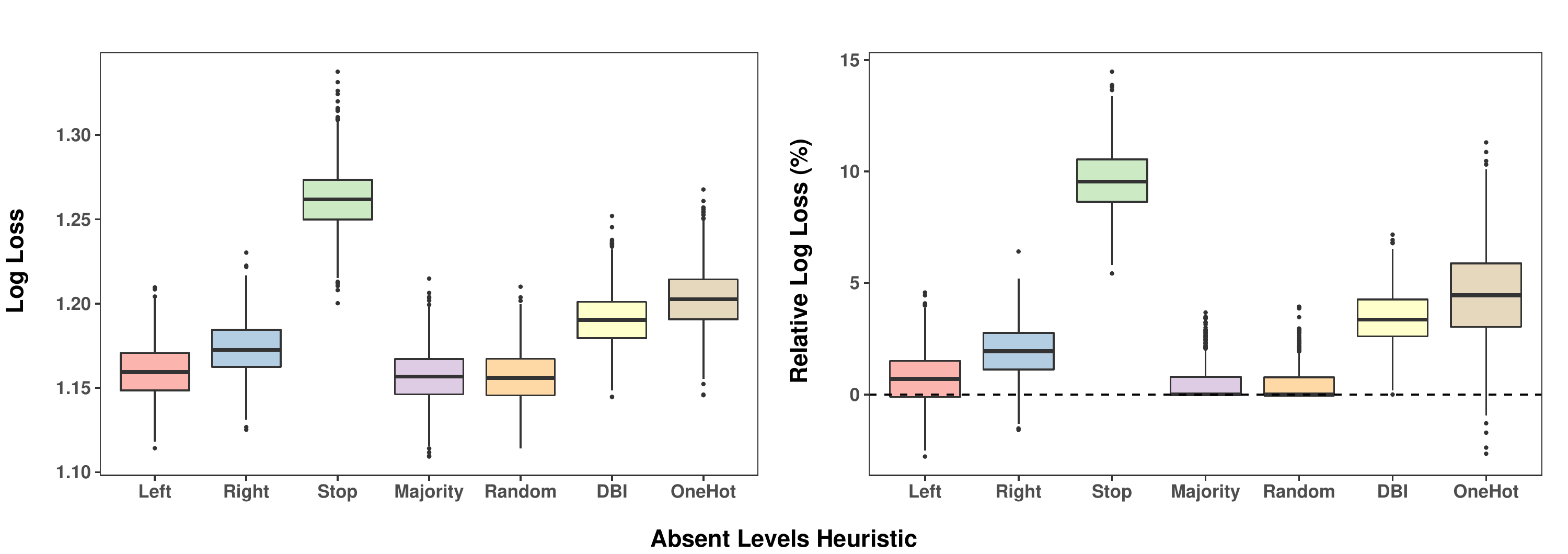}
  \captionsetup{format = hang}
  \caption{Log losses for the OOB predicted response class probabilities of the
    seven heuristics in the Pittsburgh Bridges data.  The left panel shows
    boxplots of each heuristic's marginal distribution of log losses set across
    all $1000$ experimental replications, which ignores the positive correlation
    that exists between the naive and missing data heuristics.  The right panel
    accounts for this positive correlation by comparing the log losses of the
    heuristics relative to the best log loss that was obtained amongst the
    missing data heuristics within each of the $1000$ experimental replications
    as in \eqref{eq:r_log_loss}.}
  \label{fig:bridges_loss_plot}
\end{figure}

Within each experimental replication $r$, we can use the log loss
\[
\text{LogLoss}_{r}^{(h)} = -\frac{1}{N}\sum_{n = 1}^{N}\sum_{k = 1}^{K}\!\left[I\!\left(y_{n} = k\right) \cdot log\left(\hat{p}_{nkr}^{(h)}\right)\right]
\]
to evaluate the overall performance of each heuristic $h \in \mathcal{H}$, where
we once again let $\hat{p}_{nkr}^{\left(h\right)}$ denote the OOB predicted
probability that a heuristic $h$ assigns to an observation $n$ of belonging to a
response class $k$ in an experimental replication $r$.  The left panel of
Figure~\ref{fig:bridges_loss_plot} displays the marginal distribution of each
heuristic's log losses across all $1000$ experimental replications.  However, to
once again account for the positive correlation that exists amongst the naive
and missing data heuristics, within every experimental replication $r$, we also
compare the log losses for each heuristic $h \in \mathcal{H}$ relative to the
best log loss that was achieved amongst the missing data heuristics
$\mathcal{H}_{m} = \left\{\textit{Stop}, \textit{Majority}, \textit{Random},
\textit{DBI}\,\right\}$:
\begin{equation}
  \text{LogLoss}_{r}^{\left(h \relative \mathcal{H}_{m}\right)} = \frac{\text{LogLoss}_{r}^{\left(h\right)} - \min_{h \in \mathcal{H}_{m}}\text{LogLoss}_{r}^{\left(h\right)}}{\min_{h \in \mathcal{H}_{m}}\text{LogLoss}_{r}^{\left(h\right)}}.
\label{eq:r_log_loss}
\end{equation}
Boxplots of these relative log losses are depicted in the right panel of
Figure~\ref{fig:bridges_loss_plot}.

\subsubsection{Naive Heuristics}

Although we once again stress the systematically biased nature of the Left and
Right heuristics, we note from Figure~\ref{fig:bridges_loss_plot} that the two
naive heuristics are sometimes able to outperform the missing data heuristics.
Nevertheless, on average, the Left and Right heuristics resulted in log losses
that are $0.7\%$ and $1.9\%$ worse than the best performing missing data
heuristic, respectively.

\subsubsection{Missing Data Heuristics}
Figure~\ref{fig:bridges_loss_plot} shows that for this particular example, the
Majority and Random heuristics perform roughly on par with one another, and that
they both also significantly outperform the Stop and DBI heuristics---the
smallest log loss amongst all of the missing data heuristics was achieved by
either the Majority or the Random heuristic in $999$ out of the $1000$
experimental replications.

\subsubsection{Feature Engineering Heuristic}
It can also be observed from Figure~\ref{fig:bridges_loss_plot} that, although
the One-Hot heuristic can occasionally outperform the missing data heuristics,
on average, it yields a log loss that is $4.5\%$ worse than the best performing
missing data heuristic.

\section{Conclusion}
\label{sec:conclusion}

In this paper, we introduced and investigated the absent levels problem for
decision tree based methods.  In particular, by using Breiman and Cutler's
random forests \texttt{FORTRAN} code and the \texttt{randomForest} \texttt{R}
package as motivating case studies, we showed how overlooking the absent levels
problem could systematically bias a model.  Furthermore, we presented three real
data examples which illustrated how absent levels can dramatically alter a
model's performance in practice.

Even though a comprehensive theoretical analysis of the absent levels problem
was beyond the scope of this paper, we empirically demonstrated how some simple
heuristics could be used to help mitigate the effects of absent levels.  And
although none of the missing data and feature engineering heuristics that we
considered performed uniformly better than all of the others, they were all
shown to be superior to the biased naive approaches that are currently being
employed due to oversights in the software implementations of decision tree
based methods.

Consequently, until a more robust theoretical solution is found, we encourage
the software implementations which support the native categorical split
capabilities of decision trees to incorporate the Random heuristic as a
provisional measure given its reliability---in all of our examples, the Random
heuristic was always competitive in terms of its performance.  Moreover, based
on our own personal experiences, we note that the Random heuristic was one of
the easier heuristics to implement on top of the \texttt{randomForest}
\texttt{R} package.  In the meantime, while waiting for these mitigations to
materialize, we also urge users who rely on decision tree based methods to
feature engineer their data sets when possible in order to circumvent the absent
levels problem---although our empirical results suggest that this may sometimes
be detrimental to a model's performance, we believe this to still be preferable
to the alternative of having to rely on biased approaches which do not
adequately address absent levels.

Finally, although this paper primarily focused on the absent levels problem for
random forests and a particular subset of the types of analyses in which random
forests have been used, it is important to recognize that the issue of absent
levels applies much more broadly.  For example, decision tree based methods have
also been employed for clustering, detecting outliers, imputing missing values,
and generating variable importance measures \citep{breiman:2003}---tasks which
also depend on the terminal node behavior of the observations.  In addition,
several extensions of decision tree based methods have been built on top of
software which currently overlook absent levels---such as the quantile
regression forests algorithm \citep{meinshausen:2006, meinshausen:2012} and the
infinitesimal jackknife method for estimating the variance of bagged predictors
\citep{wager:2014}, which are both implemented on top of the
\texttt{randomForest} \texttt{R} package.  Indeed, given how extensively
decision tree based methods have been used, a sizable number of these models
have almost surely been significantly and unknowingly affected by the absent
levels problem in practice---further emphasizing the need for the development of
both theory and software that accounts for this issue.

\acks{The author is extremely grateful to Art Owen for numerous valuable
  discussions and insightful comments which substantially improved this paper.
  The author would also like to thank David Chan, Robert Bell, the action
  editor, and the anonymous reviewers for their helpful feedback.  Finally, the
  author would like to thank Jim Koehler, Tim Hesterberg, Joseph Kelly, Iv{\'a}n
  D{\'i}az, Jingang Miao, and Aiyou Chen for many interesting discussions.}

\vskip 0.2in
\bibliography{16-474}

\end{document}